\documentclass[10pt,twocolumn,letterpaper]{article}
\usepackage{iccv}
\usepackage[accsupp]{axessibility}  %
\usepackage{times}
\usepackage{epsfig}

\usepackage{graphicx}

\makeatletter
\@namedef{ver@everyshi.sty}{}
\makeatother
\usepackage{tikz}
\usepackage{comment}
\usepackage{amsmath,amssymb} %
\usepackage{wrapfig} %
\usepackage{color}

\usepackage{mathtools}
\usepackage[accsupp]{axessibility}  %

\usepackage[utf8]{inputenc} %
\usepackage[T1]{fontenc}    %
\usepackage{xr-hyper}
\usepackage[pagebackref=true,breaklinks=true,letterpaper=true,colorlinks,bookmarks=false]{hyperref}
\usepackage{url}            %
\usepackage{booktabs}       %
\usepackage{xcolor,colortbl}         %
\usepackage{caption}

\usepackage{multirow}
\usepackage{enumitem}

\usepackage[normalem]{ulem}

\usepackage{subcaption}
\usepackage{float}
\usepackage{lscape}     

\usepackage{xspace}
\usepackage{setspace}
\usepackage{pifont}
\usepackage{comment}

\newif\ifblurface
\blurfacetrue
\newif\ifarxiv
\arxivtrue %

\ifarxiv
\iccvfinalcopy %
\fi

\def\iccvPaperID{4042} %

\ificcvfinal\fi

\begin{document}

\title{NPC: Neural Point Characters from Video}

\author{Shih-Yang Su$^1$ \qquad Timur Bagautdinov$^2$ \qquad Helge Rhodin$^1$\\
$^1$The University of British Columbia \quad $^2$Reality Labs Research
}

\newcommand{\TODO}[1]{\textcolor{red}{TODO: #1}}
\newcommand{\sy}[1]{{\color{blue}#1}}
\newcommand{\SY}[1]{{\color{blue}(Shih-Yang:#1)}}
\newcommand{\hr}[1]{{\color{pink}#1}}
\newcommand{\HR}[1]{{\color{pink}(Helge:#1)}}
\newcommand{\tb}[1]{{\color{cyan}#1}}
\newcommand{\TB}[1]{{\color{cyan}(Timur:#1)}}
\newcommand{\fy}[1]{{\color{green}#1}}
\newcommand{\FY}[1]{{\color{green}(Frank:#1)}}
\newcommand{\bw}[1]{{\color{green}#1}}
\newcommand{\BW}[1]{{\color{green}(Bastian:#1)}}

\newcommand{\tbf}[1]{\textbf{#1}}
\newcommand{\topic}[1]{\textbf{#1}}
\newcommand{\figref}[1]{Figure~\ref{#1}}
\newcommand{\secref}[1]{Section~\ref{#1}}
\newcommand{\feqref}[1]{Equation~\eqref{#1}}
\newcommand{\tabref}[1]{Table~\ref{#1}}

\newcommand{\app}{appendix} %

\newcommand{\parag}[1]{\noindent\textbf{#1}}

\newcommand{\new}[1]{{\color{red}{#1}}}
\newcommand{\old}[1]{\textcolor{red}{\sout{#1}}}

\newcommand{\R}{\mathbb{R}}

\newcommand{\va}{\mathbf{a}}
\newcommand{\vb}{\mathbf{b}}
\newcommand{\vc}{\mathbf{c}}
\newcommand{\vd}{\mathbf{d}}
\newcommand{\ve}{\mathbf{e}}
\newcommand{\vf}{\mathbf{f}}
\newcommand{\vg}{\mathbf{g}}
\newcommand{\vh}{\mathbf{h}}
\newcommand{\vi}{\mathbf{i}}
\newcommand{\vj}{\mathbf{j}}
\newcommand{\vk}{\mathbf{k}}
\newcommand{\vl}{\mathbf{l}}
\newcommand{\vm}{\mathbf{m}}
\newcommand{\vn}{\mathbf{n}}
\newcommand{\vo}{\mathbf{o}}
\newcommand{\vp}{\mathbf{p}}
\newcommand{\vq}{\mathbf{q}}
\newcommand{\vr}{\mathbf{r}}
\newcommand{\vt}{\mathbf{t}}
\newcommand{\vu}{\mathbf{u}}
\newcommand{\vv}{\mathbf{v}}
\newcommand{\vw}{\mathbf{w}}
\newcommand{\vx}{\mathbf{x}}
\newcommand{\vy}{\mathbf{y}}
\newcommand{\vz}{\mathbf{z}}

\newcommand{\mA}{\mathbf{A}}
\newcommand{\mB}{\mathbf{B}}
\newcommand{\mC}{\mathbf{C}}
\newcommand{\mD}{\mathbf{D}}
\newcommand{\mE}{\mathbf{E}}
\newcommand{\mF}{\mathbf{F}}
\newcommand{\mG}{\mathbf{G}}
\newcommand{\mH}{\mathbf{H}}
\newcommand{\mI}{\mathbf{I}}
\newcommand{\mJ}{\mathbf{J}}
\newcommand{\mK}{\mathbf{K}}
\newcommand{\mL}{\mathbf{L}}
\newcommand{\mM}{\mathbf{M}}
\newcommand{\mN}{\mathbf{N}}
\newcommand{\mO}{\mathbf{O}}
\newcommand{\mP}{\mathbf{P}}
\newcommand{\mQ}{\mathbf{Q}}
\newcommand{\mR}{\mathbf{R}}
\newcommand{\mS}{\mathbf{S}}
\newcommand{\mT}{\mathbf{T}}
\newcommand{\mU}{\mathbf{U}}
\newcommand{\mV}{\mathbf{V}}
\newcommand{\mW}{\mathbf{W}}
\newcommand{\mX}{\mathbf{X}}
\newcommand{\mY}{\mathbf{Y}}
\newcommand{\mZ}{\mathbf{Z}}

\newcommand{\cA}{\mathcal A}
\newcommand{\cB}{\mathcal B}
\newcommand{\cC}{\mathcal C}
\newcommand{\cD}{\mathcal D}
\newcommand{\cE}{\mathcal E}
\newcommand{\cF}{\mathcal F}
\newcommand{\cG}{\mathcal G}
\newcommand{\cH}{\mathcal H}
\newcommand{\cI}{\mathcal I}
\newcommand{\cJ}{\mathcal J}
\newcommand{\cK}{\mathcal K}
\newcommand{\cL}{\mathcal L}
\newcommand{\cM}{\mathcal M}
\newcommand{\cN}{\mathcal N}
\newcommand{\cO}{\mathcal O}
\newcommand{\cP}{\mathcal P}
\newcommand{\cQ}{\mathcal Q}
\newcommand{\cR}{\mathcal R}
\newcommand{\cS}{\mathcal S}
\newcommand{\cT}{\mathcal T}
\newcommand{\cU}{\mathcal U}
\newcommand{\cV}{\mathcal V}
\newcommand{\cW}{\mathcal W}
\newcommand{\cX}{\mathcal X}
\newcommand{\cY}{\mathcal Y}
\newcommand{\cZ}{\mathcal Z}

\newcommand{\bR}{\mathbb{R}}
\newcommand{\mx}{\mathbf{x}}
\newcommand{\mj}{\mathbf{j}}
\newcommand{\mb}{\mathbf{b}}

\definecolor{Gray}{gray}{0.85}

\definecolor{DeepGreen}{rgb}{0.15,0.60,0.15}
\newcommand{\canon}[1]{#1^c}
\newcommand{\world}[1]{#1^w}
\newcommand{\Ltwo}[1]{\vert\vert #1\vert\vert^2_2}

\newcommand{\obs}[1]{#1^\vo}

\newcommand{\ourapproach}{NPC}
\newcommand{\lbsmlp}{\text{MLP}_{LBS}}
\newcommand{\spatialmlp}{\text{MLP}_{spatial}}
\newcommand{\deformmlp}{\text{MLP}_{pose}}
\newcommand{\softmax}{\text{Softmax}}
\newcommand{\radiance}{\vc}
\newcommand{\sdf}{s}
\newcommand{\cmark}{\ding{51}}%
\newcommand{\xmark}{\ding{55}}%
\newcommand{\tmark}{\ding{115}}%

\twocolumn[{
\renewcommand\twocolumn[1][]{#1}
\maketitle
\begin{center}%
    \centering%
    \captionsetup{type=figure}%
\ifblurface
\includegraphics[width=0.82\textwidth,trim=280 310 180 290,clip]{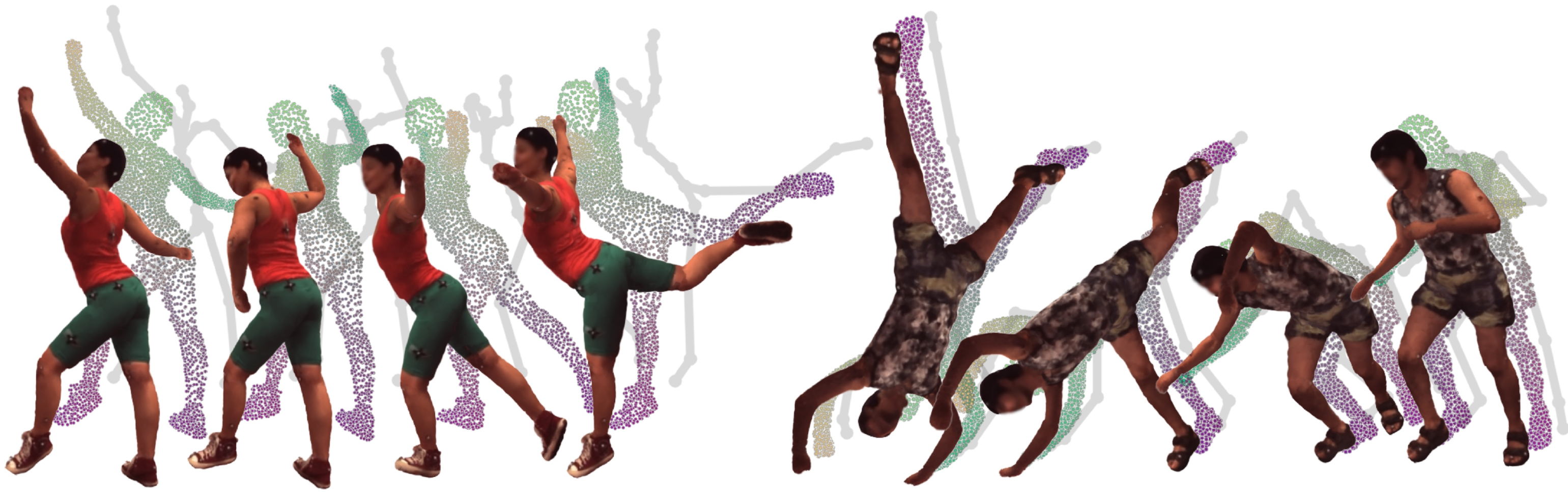}%
\else
\includegraphics[width=0.82\textwidth,trim=280 310 180 290,clip]{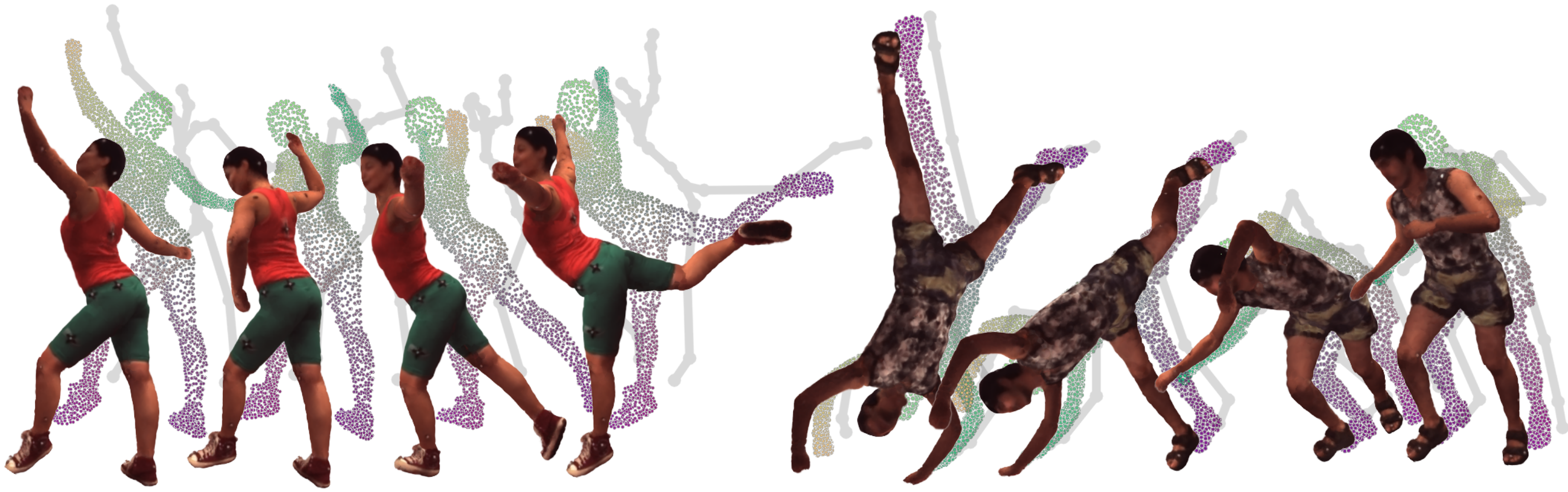}%
\fi%
    \captionof{figure}{
    \textbf{Neural Point Characters (NPC)} is an animatable point-based body model that improves fidelity and generality. 
    NPC can be learned from a single or multiple videos, generalizes well to novel poses and does not require a pre-built surface. \textbf{Back to front:} driving motion, estimated point cloud, and neural character model.%
    \ifblurface%
    {\color{red}~All faces are blurred for anonymity.}
    \else%
    \fi%
    }
\label{fig:teaser}%
\end{center}%

}]
\ificcvfinal\thispagestyle{empty}\fi
\begin{abstract}%
High-fidelity human 3D models can now be learned directly from videos, 
typically by combining a template-based surface model with neural representations.
However, obtaining a template surface requires expensive multi-view capture systems, laser scans, or strictly controlled conditions.
Previous methods avoid using a template but rely on a costly or ill-posed mapping from observation to canonical space. 
We propose a hybrid point-based representation for reconstructing animatable characters
that does not require an explicit surface model, while being generalizable to novel poses. 
For a given video, our method automatically produces an explicit
set of 3D points representing approximate canonical geometry, 
and learns an articulated deformation model that produces
pose-dependent point transformations.
The points serve both as a scaffold for high-frequency neural 
features and an anchor for efficiently mapping between observation and canonical space. 
We demonstrate on established benchmarks that our representation 
overcomes limitations of prior work operating in either canonical or in observation space.
Moreover, our automatic point extraction approach enables learning models of human and animal characters alike, matching the performance of the methods using rigged surface templates despite being more general.
\ifarxiv
Project website: {\color{magenta}\url{https://lemonatsu.github.io/npc/}}.
\fi\\
\end{abstract}\vspace{-1cm}%
\section{Introduction}
It is now possible to reconstruct photo-realistic characters from monocular videos, 
but reaching high-fidelity reconstructions still requires controlled conditions and dedicated
capture hardware that prevents large-scale use.
While static scenes can be reconstructed from multiple views recorded with a single moving camera, 
capturing dynamic human motion demands controlled studio conditions~\cite{bagautdinov2021driving,liu2021neuralactor,Lombardi21mixture,remelli2022drivable}, usually 
with a large number of synchronized cameras.
One way to tackle the monocular case is to exploit the intuition that movement of the camera with respect to a body 
part is roughly equivalent to moving a body part with respect to the camera~\cite{su2022danbo,su2021anerf}.
However, the unconstrained setting remains difficult, as it requires establishing correspondences
across frames and through dynamic deformation.

A prominent line of work follows the traditional animation pipeline---rigging a surface mesh to an underlying skeleton and equipping
the mesh either with a neural texture~\cite{bagautdinov2021driving,liu2021neuralactor} or learnable vertex 
features~\cite{kwon2021neural,peng2021animatable,peng2020neuralbody}. 
This approach is very efficient, as the forward mapping using forward kinematics provides a robust estimate of underlying geometry in closed form, and it also allows for high-quality reconstructions, as neural textures are capable of
representing high-frequency details~\cite{liu2021neuralactor,thies2019neuraltex}. 
However, it does require highly accurate 3D pose and body shape, which are typically obtained from expensive laser 
scans~\cite{wang2022arah} or by offline fitting personalized parametric body models to multi-view captures~\cite{bagautdinov2021driving,xu2021hnerf}. 
Moreover, most of the existing parametric models rely on linear blend skinning~\cite{lewis2000pose}, which is prone to artefacts.
In this work, we aim at building an animatable full-body model from a single monocular video, without 
relying on a pre-defined template or complex capture systems. 

Another line of work reduces the problem to building a body model in canonical space, 
e.g., learning a neural radiance field of the person in T-pose~\cite{jiang2022neuman,li2022tava,wang2022arah}.
In practice, this family of approaches requires finding the backward mapping
from the observation space to the canonical space, which is either learned from high-fidelity scans~\cite{wang2021metaavatar},
e.g. by learning a deformation on top of rigid motion~\cite{noguchi2022unsupervised,weng2022humannerf}, 
or through root finding~\cite{li2022tava,wang2022arah}.
These methods are typically limited in generalization, require high-quality training data, and are computationally heavy
at test time.

Our goal is to attain the generality of surface-free canonical models and the speed of forward
mappings without a need for a pre-defined template that restricts applicability. 
To this end, we rely on an implicit, surface-free body representation that does not depend on a 
precise 3D pose nor a rig~\cite{su2022danbo,su2021anerf}. 
Namely, we use A-NeRF~\cite{su2021anerf} alongside off-the-shelf pose estimators~\cite{kolotouros2019learning_spin} to get a reasonable 
3D pose estimate and DANBO~\cite{su2022danbo} 
to learn an initial surface estimate in the canonical space. %

Given such an initial estimate, our method, Neural Point Characters (NPC), reconstructs a high-quality neural 
character that can be animated with new poses and rendered in novel views (\figref{fig:teaser}). 
The difficulty is that the initial shape is very approximate and noisy, and is
insufficient to model high-quality geometry.
Central to NPC is thus a novel point representation that is designed to improve noisy shape 
estimates, and subsequently learns to represent fine texture details and pose-dependent deformations. 
Keys are our two main contributions:
First, we propose to find correspondences between the canonical and observation space by inverting the forward mapping %
via nearest-neighbor lookup, including the non-linear pose-dependent deformation field modeled with a graph neural network (GNN).
The surface points serve as anchors for inferring the backward mapping of a query point to the neural field in canonical 
space through an efficient amortized lookup. 
Our approach is more efficient than root finding and more precise than models assuming piece-wise rigidity. 
Second, we propose to use the non-linearly deformed points as a scaffold in observation space and represent high-frequency 
pose-dependent details. Point features and neural fields in canonical space are further combined with 
bone-relative geometry encodings to encourage better generalization.
We provide a comprehensive evaluation of our method, and demonstrate state-of-the-art results on the established 
benchmarks in monocular and multi-view human reconstruction. 
We additionally demonstrate versatility by reconstructing human and animal characters alike.

\section{Related Work}
\paragraph{Neural rendering} has recently emerged as a powerful family of approaches for 
building controllable neural representations of 3D scenes~\cite{meshry2019neuralrerendering,mildenhall2020nerf,tewari2020neuralrendering,wiles2020synsin}. 

\parag{Point-Based Representations.}
NPBG~\cite{aliev2019neuralpoint} uses a pre-computed static point cloud to represent a rough scene geometry, and learns a set of neural descriptors attached to each point, which are then rasterized 
at multiple resolutions and further processed with a 
learnable rendering network.
ADOP~\cite{ruckert2022adop} follows a similar strategy but
additionally optimizes camera parameters and point cloud locations, 
which leads to better fidelity reconstructions.
Pulsar~\cite{lassner2020pulsar} represents a scene as a collection of
spheres, which have optimizable locations and radius parameters and are then rendered 
with a custom differentiable renderer and post-processed with neural shading ConvNet.
SMPLpix~\cite{prokudin2021smplpix} and Point-based human clothing~\cite{zakharkin2021point} %
similarly utilize a 2D CNN to in-paint the rasterized 2D %
point feature maps for rendering. %
Power of Points~\cite{ma2021power} learns an auto-decoder to predict point displacements for modeling different garments and dynamics. Concurrent work PointAvatar~\cite{Zheng2023pointavatar} represents facial avatars with deformable point clouds that grow progressively to capture fine-grained details, and render with a differential point rasterizer.
Our method also builds upon a point-based representation, but is equipped with an articulated model and relies on volumetric rendering.

\parag{Neural Field Representations.} 
Neural Radiance Fields (NeRF)~\cite{mildenhall2020nerf,sitzmann2020implicit_siren} represent scenes implicitly, 
as a coordinate-conditioned MLPs and enable high-fidelity novel-view synthesis of static scenes.
NeRF-like methods have also been applied to dynamic scenes by adding a deformation field~\cite{gao2021dynerf,li2021neural,park2021hypernerf,pumarola2020dnerf,xian2021space}. 
Particularly relevant to our method is Point-NeRF~\cite{xu2022point}. 
Point-NeRF uses point clouds obtained from multi-view stereo 
as a geometry proxy, and models appearance on top with a locally-conditioned 
NeRF to achieve faster learning and enables high-fidelity novel-view synthesis
on static scenes.
Similarly,~\ourapproach~ uses point-based neural fields, 
but uses pose-driven articulated skeleton model to transform 
the point cloud to enable dynamic human modeling, and 
dynamically decodes pose-dependent per-point payload to account
for pose-dependent deformations.

\parag{Neural Fields for Human Modeling.} 
Recent work applies NeRF to reconstruct dynamic human appearance~\cite{kwon2021neural,mihajlovic2022keypointnerf,peng2020neuralbody,pumarola2020dnerf,weng2022humannerf} 
and learning animatable human models~\cite{li2022tava,liu2021neuralactor,noguchi2021narf,peng2021animatable,su2022danbo,su2021anerf,wang2022arah,zheng2022structured} 
from video sequences. 
NeRF for a \textit{static} scene directly maps query point coordinates to opacity and color values.
The key challenge for an \textit{animatable} NeRF model is to map a per-frame 3D query
to the canonical body model space.

A common approach to tackle this challenge is to learn a \textbf{backward mapping} 
that transports query points from observation space with the posed person 
to the canonical space
with the person in rest pose (e.g., T-pose)~\cite{deng2020nasa,jiang2022neuman,liu2021neuralactor,mihajlovic2022compositional_coap,peng2021animatable,saito2021scanimate,tiwari21neuralgif}. 
The rendering is then done by neural fields defined in the canonical space.
While learning backward mapping allows for high-quality reconstructions of dynamic human geometry and appearance, 
these methods typically require
3D scans for training~\cite{deng2020nasa,mihajlovic2022compositional_coap,saito2021scanimate,tiwari21neuralgif}, 
need a texture map~\cite{liu2021neuralactor}, or depend on a surface prior~\cite{jiang2022neuman,peng2021animatable}. 
Additionally, the backward mapping is difficult to learn as it requires solving many-to-one mapping~\cite{chen2021snarf,li2022tava}. These approaches are thus prone to artefacts when rendering unseen poses. 

Alternatively, some existing methods rely on a \textbf{forward mapping} 
that moves features in the canonical space to the observation space. 
Earlier approaches anchor neural features to SMPL vertices~\cite{kwon2021neural,peng2020neuralbody} 
and deform the body mesh to the observation space via linear blend skinning (LBS)~\cite{lewis2000pose}.
Per-vertex features are then diffused by a 3D CNN around the posed mesh, 
and the neural field is defined on top of the resulting voxel feature grid.
These methods are still limited by the quality of the underlying LBS
model and inability to capture non-linear deformations, which often results in blurry reconstructions.

An alternative backward mapping that was recently popularized is to apply a differentiable root-finding
algorithm~\cite{li2022tava,wang2022arah} on learned forward mappings.
TAVA~\cite{li2022tava} is a template-free approach that performs iterative root-finding 
to solve for a point in canonical space. 
ARAH~\cite{wang2022arah} proposes a joint root-finding method that finds ray intersection on the NeRF body 
model initialized using a pre-trained hypernetwork~\cite{ha2017hypernetworks,wang2021metaavatar} 
for rendering. 
Although TAVA and ARAH are template-free and show state-of-the-art synthesis quality on unseen poses, both methods are computationally heavy
at train and test time, and ARAH additionally requires a surface
prior which is built using 3D scans. 
Compared to these approaches, \ourapproach~uses sparse point clouds to create 
efficient forward mapping between a canonical and observation space, which
provides better query feature localization with lower computational cost, 
and does not rely on pre-trained surface priors. %

Conceptually related to our method are SLRF~\cite{zheng2022structured}, KeypointNeRF~\cite{mihajlovic2022keypointnerf}, and AutoAvatar~\cite{bai2022autoavatar}. SLRF shares a similar concept of leveraging surface points for anchoring separate radiance fields to represent local appearances. In contrast, NPC explicitly represents local structure using much denser feature point clouds and enabling dense correspondence across body poses without relying on pre-defined parametric meshes. KeypointNeRF uses sparse key points to triangulate 2D pixel-aligned features extracted from multi-view imagery for 3D face reconstructions. 
Unlike KeypointNeRF, NPC stores point features directly in 3D space, captures pose-dependent deformation, and is drivable. AutoAvatar uses K-NN-based encodings for dynamic neural body deformation fields, but does not model textures.

\begin{figure*}[t]
\centering
\ifblurface
\includegraphics[width=0.878\linewidth,trim=0 5 0 5,clip]{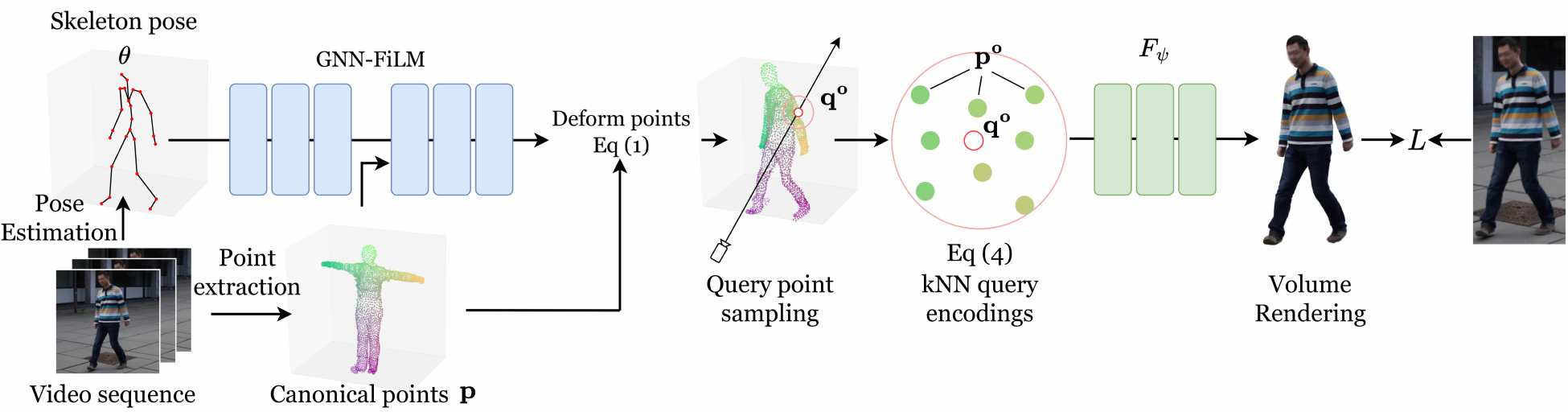}%
\else
\includegraphics[width=0.878\linewidth,trim=0 5 0 5,clip]{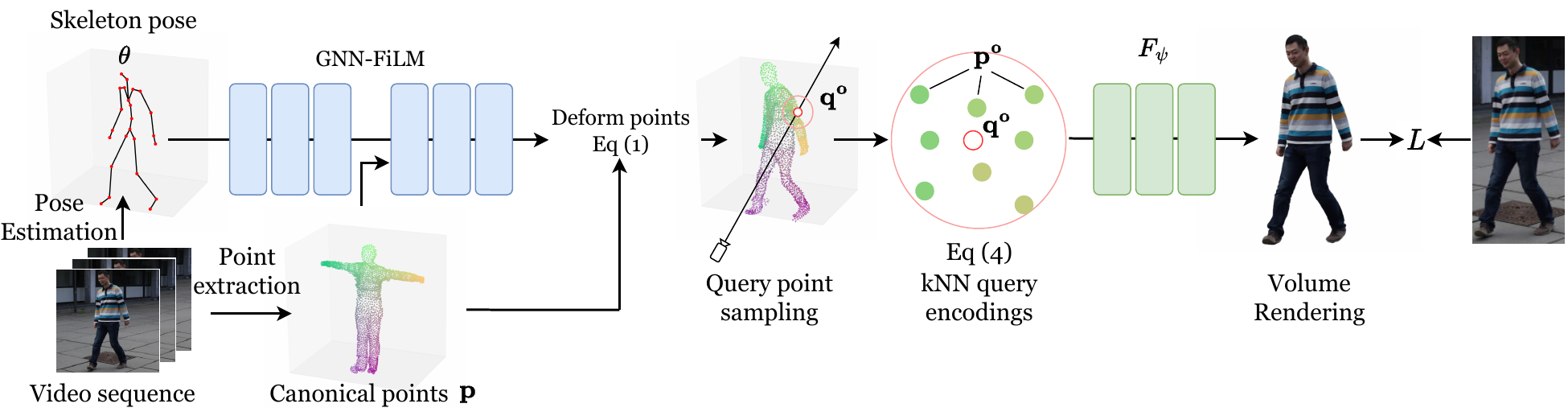}%
\fi
\caption{\textbf{Overview.} 
NPC produces a volume rendering of a character with a NeRF $F_\psi$ locally conditioned on 
features aggregated from a dynamically deformed point cloud. 
Given a raw video, we first estimate a canonical point cloud $\vp$ with an implicit body model (\secref{sec:method-initialization}). 
GNN then deforms canonical points $\vp$ conditioned on skeleton pose $\theta$, 
and produces a set of pose-dependent per-point features (\secref{sec:method-transformation}, \secref{sec:method-feature}).
Every 3D query point $\vq^\vo$ in the observation space aggregates the features 
from k-nearest neighbors in the posed point cloud. %
The aggregated feature is passed into $F_\psi$ for the volume rendering.
Our model is supervised directly with input videos (\secref{sec:training}). 
}%
\vspace{-3mm}
\label{fig:overview}
\end{figure*}

\section{Method}
\begin{figure*}[h]
\centering
\includegraphics[width=0.765\linewidth,trim=10 5 80 20,clip]{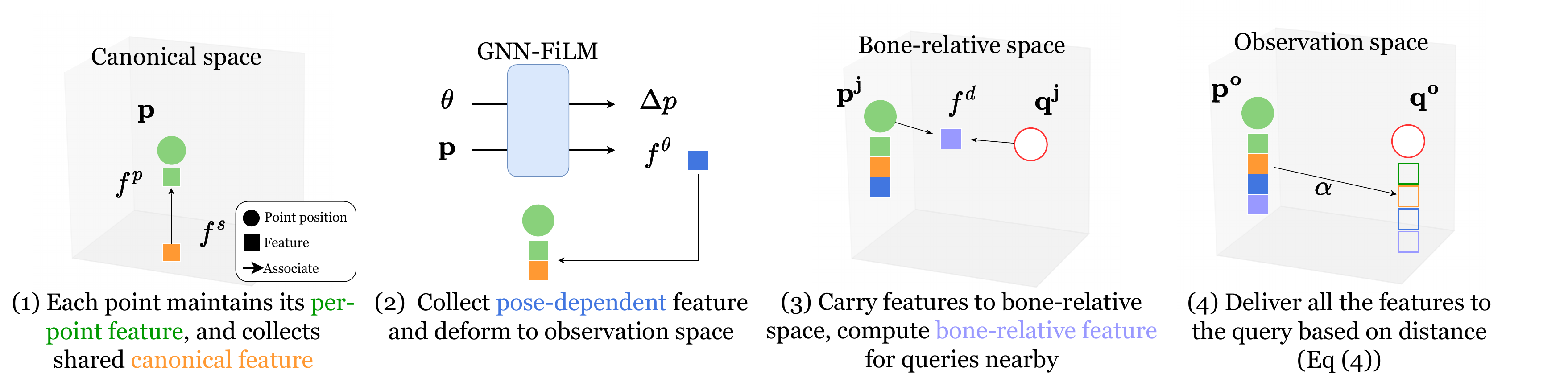}%
\caption{\textbf{Point feature encoding.} 
Our core idea is to employ a point cloud $\vp$ as an anchor 
to carry features from the canonical to the observation space, forming an efficient mapping between the two. (1) %
Each $\vp$ carries a learnable feature $f^p$ and its position queries features $f^s$ from 
a canonical field. 
(2) The GNN adds pose-dependent features $f^\theta$ and deformation $\Delta \vp$. 
(3) The view direction and distance is added in bone-relative space. 
(4) The k-nearest neighbors of $q^\vo$ in $\{\vp^\vo_i\}_i^N$ are used to establish forward and backward mapping from a query point to both posed and canonical points.
}
\vspace{-3mm}
\label{fig:point-feature}
\end{figure*}

NPC reconstructs a neural character that can be animated 
from a single video 
$\{\mI_1,\cdots,\mI_{N}\}$ with $N$ frames. 
\figref{fig:overview} provides an overview. 
The representation includes neural fields~\cite{mildenhall2020nerf} in canonical space 
with the character in rest pose and a set of surface points $\{\vp_1,\cdots,\vp_{P}\}$ 
that deform dynamically and map between canonical and observation space. 
In the following, we explain how we obtain these representations and how we combine them to 
form the geometry and appearance of the NPC.

\subsection{Skeleton and Point Representation.}
\label{sec:method-initialization}
\parag{Pose and shape initialization.}
Each frame $\mI_t$ of the input video is processed with an off-the-shelf pose estimator\footnote{We follow recent template-free human NeRFs~\cite{li2022tava,su2022danbo} in using SMPL-based pose estimators to acquire joint angle rotations, but other pose estimators would work equivalently.} 
followed by refinement with A-NeRF~\cite{su2021anerf} or if available, multi-view triangulation.
The resulting pose $\theta_t = \{\mR_{t,0},\cdots,\mR_{t,J-1}\}\in\R^{J\times3}$, the joint angle rotations 
of a pre-defined skeleton with $J$ joints, is then used as input to train the existing neural body model 
DANBO~\cite{su2022danbo}. 
This body model is set to a T-pose, and the iso-surface of the underlying implicit model is extracted 
with marching cubes.
Note that we train DANBO with a much-reduced number of ray samples and 
small number (10k) of iterations, which corresponds to $\sim$2.5\% of the full training set; 
this process runs under 30 mins on a single NVIDIA RTX 3080 GPU.

\parag{Part-wise point cloud sampling.}
We obtain the location of the sparse surface points $\vp$ by farthest point sampling on the 
isosurface until we obtain 200 points $\{\vp_{i,j}\}_{i=1}^{200}$ for every body part $j$ of the skeleton. 
We drop the part subscript for simplicity when the part assignment is irrelevant. 
As demonstrated in Figure~\ref{fig:overview}, the resulting point cloud is a very coarse
approximation of the underlying geometry.
In what follows, we describe our main contributions, which allow learning
high-quality details on top of this rough estimate, without strict assumptions 
on a clean surface imposed by existing model-based approaches, and with 
sharper details and better texture consistency compared to DANBO and other surface-free models.

\parag{Canonical to observation space (forward mapping).}
\label{sec:pts-deformation}
The extracted points $\vp_i\in \R^3$ live in canonical space and are fixed for the 
entire video. 
Deformations in frame $t$ are modelled by a learnable mapping $\vp_{i,j}$ to the posed observation space,
\begin{equation}
    \obs{\vp_{i,j,t}} = \text{LBS}(\theta_t, \vw, \vp_{i,j}) + R_{t,j}(\theta_t)\Delta \vp_{i,j}(\theta_t),
    \label{eq:deform-full}
\end{equation}
comprising a linear blend skinning (LBS)~\cite{lewis2000pose} term to model rigid deformations 
and a non-linear term $\Delta \vp(\theta_t)$ that is learned relatively to the orientation $R_{t,j}(\theta_t)$ 
of the underlying skeleton.

Key to our approach is learning the pose-dependent deformation $\Delta \vp_{i,j}(\theta_t)$ and 
character-specific LBS weights $\vw$ that define the influence of every joint on every point. 
This is distinct from prior work using template meshes with artist-created LBS weights~\cite{liu2021neuralactor,peng2021animatable,peng2020neuralbody,wang2022arah}.
We initialize the learnable LBS weights for a point $\vp$ to
\begin{equation}
    \vw^0=\left\{w_1,\cdots,w_j\right\},\text{~where~} w_j=\frac{1}{\sqrt{d_j(\vp)}},
\end{equation}
with $d_j(\cdot)$ being the distance from point $\vp$ to the bone $j$. 
In the following, we drop the frame index $t$ when irrelevant.

\subsection{Neural Rendering with a Point Scaffold}
\label{sec:method-transformation}
An image $\hat{\mI}$ of the character is created by volume rendering of an implicit
radiance and signed-distance field $(\radiance, \sdf)$ as in~\cite{yariv2021volume}. 
The rendered image are used for optimizing NPC as detailed in  Section~\ref{sec:training}. 
Specific to our method is how we combine various point features that are sparse to form a continuous field $f(\vq^\vo)$.
Given a body pose $\theta$ and a 3D query point $\vq^\vo$ in the observation space, 
we predict the corresponding color $\radiance$ and signed distance $\sdf$ with a 
neural field $F_\psi$ with learnable parameter $\psi$ from $f(\vq^\vo)$, 
combined with established frame codes $f(t)$~\cite{martin2020nerfiw,peng2020neuralbody} 
and geometric features $g(\vq^\vo)$,
\begin{equation}
    (\radiance, \sdf) = F_\psi(f(\vq^\vo),g(\vq^\vo),f(t),\vw) \; .
    \label{eq:nerf-predict}
\end{equation}
The individual features are introduced in the subsequent section and supplemental document. 
We continue with the sparse-to-continuous transformations and mapping between observation and canonical spaces. 

\parag{Point features to observation space (forward mapping).} Our per-point features encode the local appearance and shape. 
They are associated with points $\vp_i$, living in canonical space. 
To map features from canonical to observation space we first apply Eq.~\ref{eq:deform-full}, 
yielding 3D locations $\vp^\vo$. 
To create a continuous field, we query the $K=8$ nearest neighbors of $\vq^\vo$ and aggregate features $f_i$ with a Gaussian RBF kernel,%
\begin{equation}
    f(\vq^\vo) = \sum_{i\in \cN(\vq^\vo)} f_i a_i,\quad a_i = \exp\left(- \frac{(\vq^\vo-\vp^\vo_i)^2}{\beta_i}\right) \; ,
    \label{eq:accumulation}
\end{equation}
where $\beta_j$ is a learnable scale parameter that determines the influence range of feature $i$ among nearest neighbors $\cN(\vq^\vo)$.

\parag{Observation to bone space (rigid backward mapping).} 
The kinematic chain of the skeleton in pose $\theta$ defines for every joint $j$ a rigid transformation $\mT_j$ and its inverse,
\begin{equation}
    \vq^j = \mT^{-1}_j(\theta) \vq^\vo,
\end{equation}
that maps from query $\vq^\vo$ in posed to joint-relative space. 
It is efficient and widely used but neglects non-rigid motion.

\parag{Observation to canonical space (non-rigid backward).} 
Inverting the non-linear pose-dependent deformation is a non-trivial task. 
Prior work tried to approximate it with another network~\cite{noguchi2021narf,peng2021animatable,tiwari21neuralgif}, 
but this requires conditioning on the position in posed space 
and leads to poor generalization.
Another approach is iterative root finding~\cite{li2022tava,wang2022arah}, which is
extremely slow, both at training and at inference time. 

Instead, we exploit the fact that the forward mapping of surface points via Eq.~\ref{eq:deform-full} 
lets us invert the function point-wise by using points as the anchor. 
The forward mapping gives a finite number of pairs $(\vp^\vo,\vp)$ in observed and canonical space that correspond exactly.
Moreover, the features important for shape reconstruction live on the surface, and so 
we found it sufficient to query the canonical space feature $f^s$ at the canonical 
positions of the nearest neighbors $\cN(\vq^\vo)$. 
Intuitively, our surface points $\vp$ serve as anchors to carry the appearance information from 
the canonical space to the point $\vq^\vo$ that resides in the observation space.

\parag{Efficient nearest neighbor lookup}
Our forward and backward mappings both rely on nearest neighbor search, 
which would be slow when performed naively over the entire point cloud for each query.
To make this approach feasible, we apply divide and conquer by searching only within the points 
associated with the three closest joints.
Moreover, because our points form a surface, we found i) that the nearest neighbors to a query stays mostly unchanged in both observation and canonical space and ii) that the one nearest neighbor defines the remaining $K-1$.
This makes it possible to tabulate for each canonical point its $K-1$ nearest neighbors.
This reduces the runtime computation to finding the one nearest neighbor in posed space and 
looking up the remaining ones. 
Therefore, we can use a large number of neighbors without runtime drawbacks. 
Note that we allow the tabulated nearest neighbors to cross joint boundaries to prevent seam artefacts.

\subsection{Neural Features and Field Representations}
\label{sec:method-feature}
With the mapping between different spaces established through our point features, we now turn to the structure of the neural features in each space and their optimization. 

\parag{Point features.}
We associate each point $i$ with a feature vector $f^p$ to provide an explicit representation local 
to the corresponding body region.
It improves texture consistency across poses and captures local high-frequency appearance and geometry.
Feature vector $f^p$ and position $\vp_i$ are both learned,
initialized with respectively $\mathcal{N}(0, 0.1)$ and the canonical point clouds from DANBO.

\parag{Pose-dependent point features.} 
We use a pose-conditioned GNN to predict both the per-point pose-dependent feature $f^\theta$ and deformation $\Delta p$,
\begin{equation}
    f^\theta,\Delta p = \text{GNN-FiLM}(\vp, \theta) \; .
\end{equation}
The architecture is inspired by DANBO~\cite{su2022danbo}, and extended with FiLM conditioning~\cite{perez2018film}.

\parag{Canonical features.}
We represent $f^s$ using a per-part factorized volume $\vv_j\in\mR^{12\times3V}$~\cite{su2022danbo}. 
To retrieve the feature for a point $\vp$, we project and interpolate on the factorized vector of each axis,
\begin{align}
\vv_j(\vp)&=\left(v^x[x(\vp)],v^y[y(\vp)],v^z[z(\vp)]\right)\\
    f^s &= \spatialmlp\left(\vv_j(\vp)\right)\in\mR^{3V} \; ,
\label{method:point-spatial-feat}
\end{align}
where $v^{(\cdot)}\left[\cdot\right]$ is the interpolated feature, and $\spatialmlp$ is a 2-layer 
MLP combining features from all axis. 

\parag{Bone-relative features.}
To provide the NeRF ability to reason about the approximation made 
in the non-linear backward mapping, we condition it on
$f^d$ - the offset of the query to each of the nearest neighbors, 
defined in the local bone coordinates of the closest joint $j$. 
This distance is computed in a learned feature space $\vc_j\in\mR^{16\times3C}$,
\begin{equation}
    f^d_i = \vc_{j}(\mR_{j}^{-1}\vp^\vo_{i,j}) - \vc_j(\mR_{j}^{-1} \vq^\vo_{j}),
\end{equation}
with $\vc_j$ a per-part factorized feature volume, like $\vv$.

After the NN lookup, all of the introduced features are associated with points 
$\vp_i$ (see~\figref{fig:point-feature}). 
Together, they form the $f_i = (f^p_i, f^\theta_i, f^s_i, f^d_i)$ used in Eq.~\ref{eq:accumulation}, 
which carries both the shared and independent traits, as well as the flexibility to adapt to 
deformation-induced appearance changes. 
Additional commonly-used features and $g(\vq)$ are explained in the supplemental document.

\subsection{Training}
\label{sec:training}
\newlength\hmnviewscale
\setlength\hmnviewscale{0.129\textwidth} %
\newcommand{\hmnviewpath}{statics/figs/exp/h36m_novelview/}

\ifblurface
\newcommand{\hmviewpost}{_blur}
\else
\newcommand{\hmviewpost}{}
\fi
\newcommand{\hmnviewma}{ARAH}
\newcommand{\hmnviewmb}{DANBO}
\newcommand{\hmnviewmc}{TAVA}
\newcommand{\hmnviewmd}{Ours-full}
\newcommand{\hmnviewa}{S7}
\newcommand{\hmnviewanum}{00}
\newcommand{\hmnviewb}{S8}
\newcommand{\hmnviewbnum}{00}
\begin{figure*}[t]
\setlength{\fboxsep}{0pt}%
\setlength{\fboxrule}{0pt}%
\parbox[t]{\hmnviewscale}{%
\centering%
\fbox{\includegraphics%
[width=\hmnviewscale]
{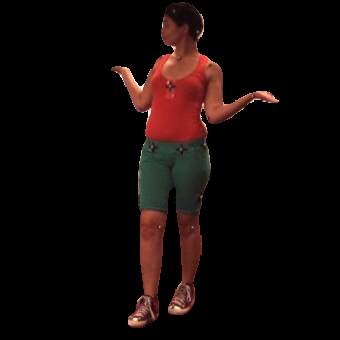}%
}\\%
\fbox{\includegraphics%
[width=\hmnviewscale]
{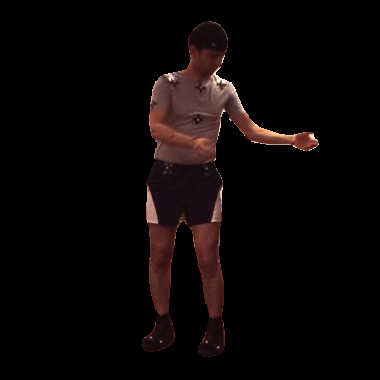}%
}\\%
{ Training view}%
}%
\hfill%
\parbox[t]{\hmnviewscale}{%
\centering%
\fbox{\includegraphics%
[width=\hmnviewscale]
{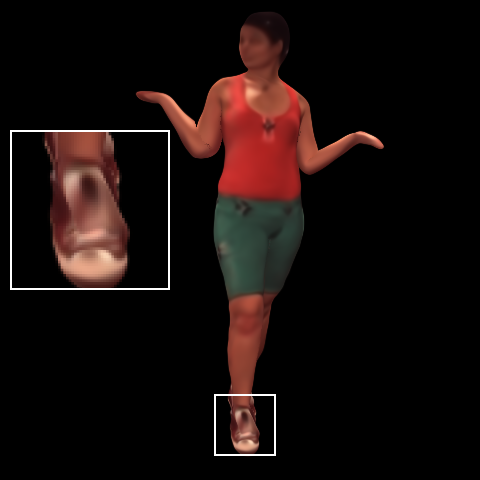}%
}\\%
\fbox{\includegraphics%
[width=\hmnviewscale]
{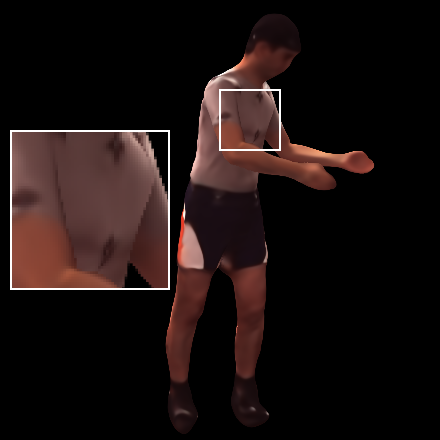}%
}\\%
{\hmnviewma}%
}%
\hfill%
\parbox[t]{\hmnviewscale}{%
\centering%
\fbox{\includegraphics%
[width=\hmnviewscale]
{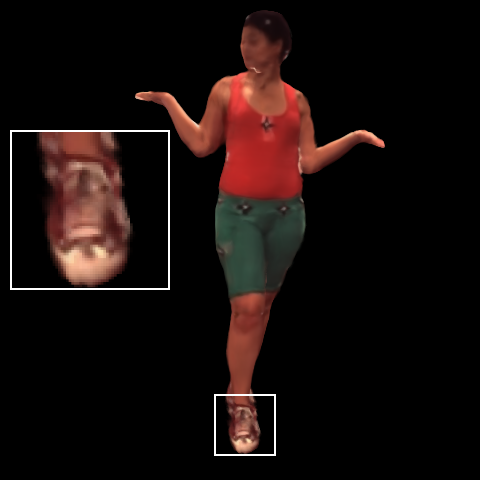}%
}\\%
\fbox{\includegraphics%
[width=\hmnviewscale]
{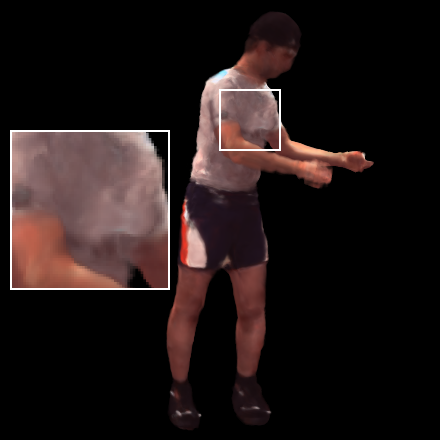}%
}\\%
{\hmnviewmb}%
}%
\hfill
\parbox[t]{\hmnviewscale}{%
\centering%
\fbox{\includegraphics%
[width=\hmnviewscale]
{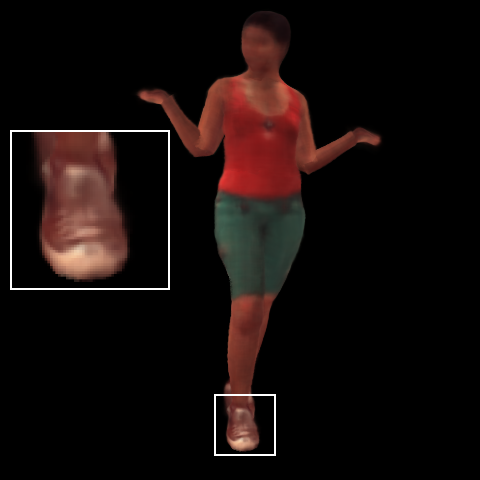}%
}\\%
\fbox{\includegraphics%
[width=\hmnviewscale]
{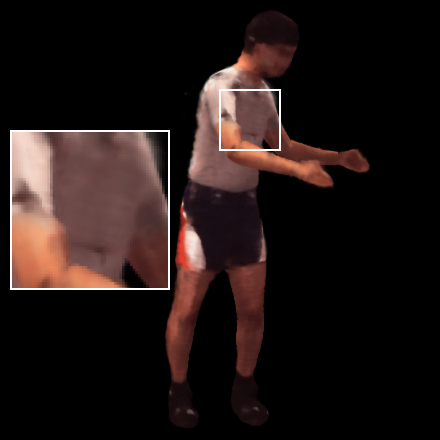}%
}\\%
{\hmnviewmc}%
}%
\hfill
\parbox[t]{\hmnviewscale}{%
\centering%
\fbox{\includegraphics%
[width=\hmnviewscale]
{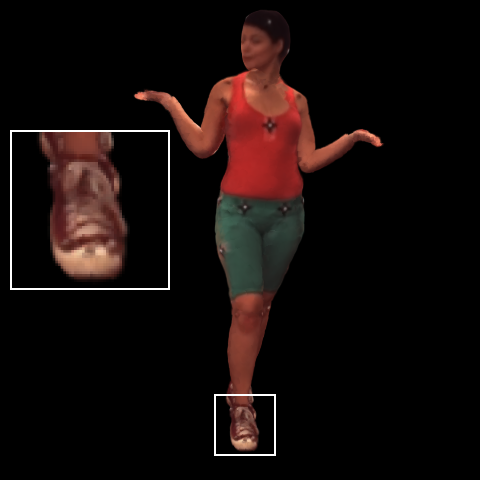}%
}\\%
\fbox{\includegraphics%
[width=\hmnviewscale]
{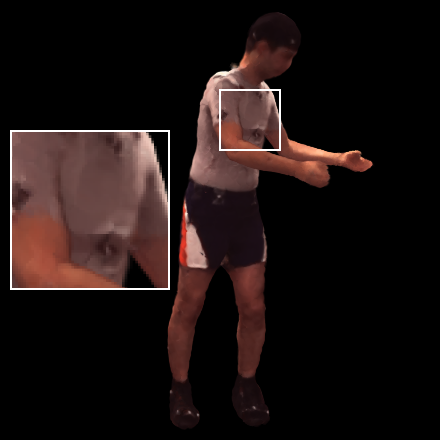}%
}\\%
{NPC (Ours)}%
}%
\hfill
\parbox[t]{\hmnviewscale}{%
\centering%
\fbox{\includegraphics%
[width=\hmnviewscale]
{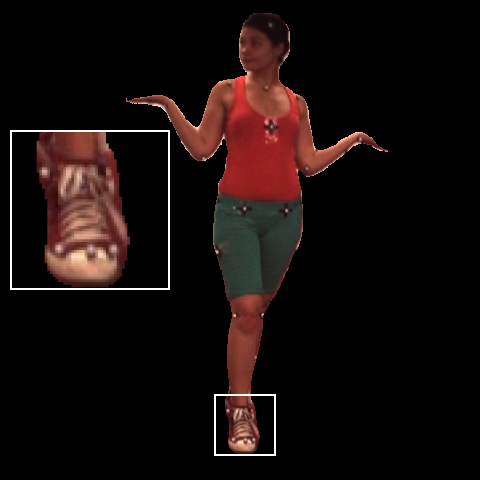}%
}\\%
\fbox{\includegraphics%
[width=\hmnviewscale]
{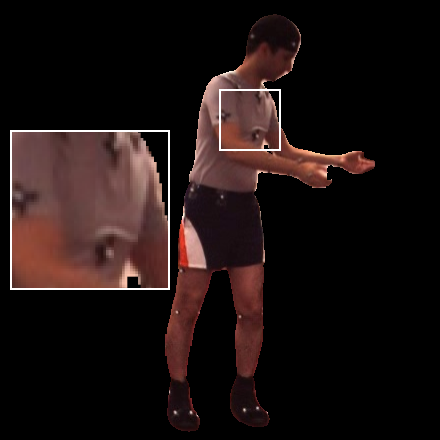}%
}\\%
{ Ground truth}%
}%
\centering%
\caption{\textbf{Novel view synthesis on Anim-NeRF Human3.6M~\cite{Ionescu11,Ionescu14a,peng2021animatable} test split}. Our point-based representation ensures better feature consistency across training poses, and therefore synthesizes better details for novel views.}%
\label{fig:exp-qualitative-novel-view}
\end{figure*}
We train~\ourapproach~with the photometric loss on ground truth images~$\mI$,
\begin{equation}
    L_p = \sum_{(u,v)\in I} \vert I(u,v,\psi) - \hat{I}(u,v)\vert.
\end{equation}
We employ L1 loss on real-world datasets for robustness and L2 otherwise.
To encourage the network to predict proper level-set, we adopt the eikonal loss~\cite{amos2020implicit,yariv2021volume} 
\begin{equation}
    L_{eik} = \sum_{\tilde{\vp}} (\vert\vert\nabla\sdf\vert\vert-1)^2 \; ,
\end{equation}
where $\tilde{\vp}$ are sparse points randomly selected around the posed surface points $\vp^\vo$. 
We regularize the output of the deformation network with
\begin{equation}
    L_{\Delta p} = \max(\Delta p - \delta, 0)^2,
\end{equation}
which penalizes the non-linear deformation when $\Delta p$ is over the threshold $\delta=0.04$. A second loss discourages the network from moving points away from their neighbors,
\begin{equation}
    L_{N}=\sum_{\vp^\vo}\sum_{\vp_i\in\cN(\vp)} (\vert\vert \vp^{\vo} - \hat{\vp}_i^\vo\vert\vert - \vert\vert \vp - \hat{\vp}_i\vert\vert)^2.
\end{equation}
We include a third loss to encourage the predicted signed distance on the surface points to either be on or within the body surface,
\begin{equation}
    L_s = \sum_{\tilde{\vp}}\max(s, 0).
\end{equation}
To summarize, our full training objective is
\begin{equation}
    L=L_p + \lambda_{eik}L_{eik} + \lambda_{\Delta p} L_{\Delta p} + \lambda_{N} L_N + \lambda_{S} L_S,
\end{equation}
with $\lambda_{eik}=0.01$, $\lambda_{\Delta p}=1.0$, $\lambda_N=10.0$, and $\lambda_{S}=0.1$.
We train for 150k iterations, for 12 hours on a single NVIDIA RTX 3080 GPU including the point initialization process.
\newlength\hmqualscale
\setlength\hmqualscale{0.129\textwidth} %
\newcommand{\hmqualpath}{statics/figs/exp/h36m_novelpose/}
\ifblurface
\newcommand{\hmqualpost}{_blur}
\else
\newcommand{\hmqualpost}{}
\fi
\newcommand{\hmqualma}{NeuralBody}
\newcommand{\hmqualmb}{ARAH}
\newcommand{\hmqualmc}{DANBO}
\newcommand{\hmqualmd}{TAVA}
\newcommand{\hmqualme}{Ours-full}
\newcommand{\hmquala}{S6}
\newcommand{\hmqualanum}{00}
\newcommand{\hmqualb}{S5}
\newcommand{\hmqualbnum}{00}
\newcommand{\hmqualc}{S9}
\newcommand{\hmqualcnum}{00}
\newcommand{\hmquald}{S11}
\newcommand{\hmqualdnum}{00}
\begin{figure*}[t]
\setlength{\fboxsep}{0pt}%
\setlength{\fboxrule}{0pt}%
\parbox[t]{\hmqualscale}{%
\centering%
\fbox{\includegraphics%
[width=\hmqualscale]
{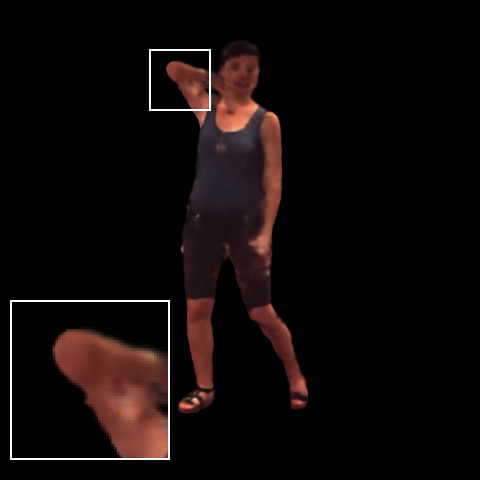}%
}\\%
\fbox{\includegraphics%
[width=\hmqualscale]
{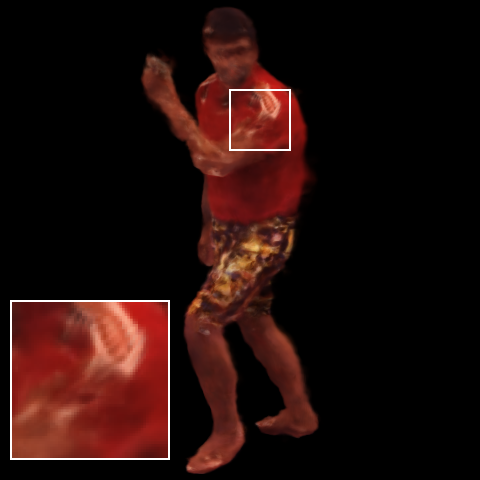}%
}\\%
\fbox{\includegraphics%
[width=\hmqualscale]
{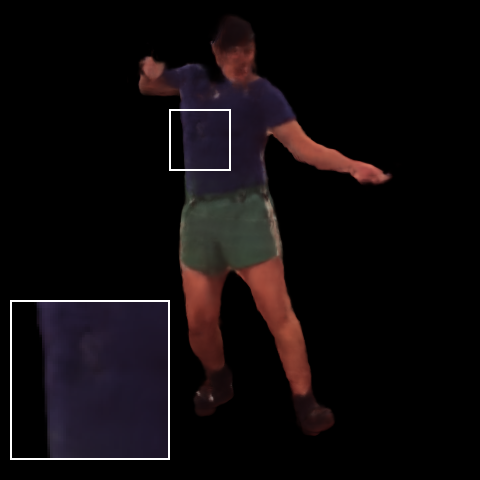}%
}\\%
{ \hmqualma}%
}%
\hfill%
\parbox[t]{\hmqualscale}{%
\centering%
\fbox{\includegraphics%
[width=\hmqualscale]
{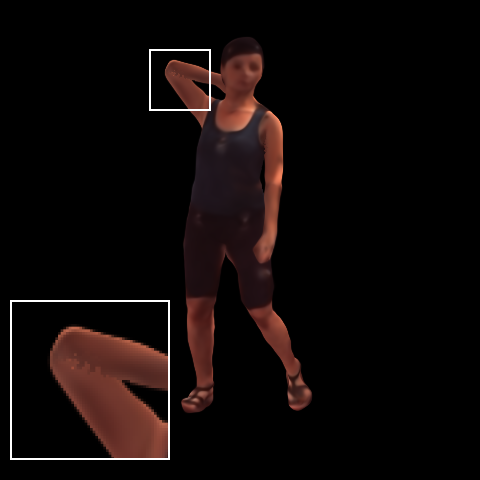}%
}\\%
\fbox{\includegraphics%
[width=\hmqualscale]
{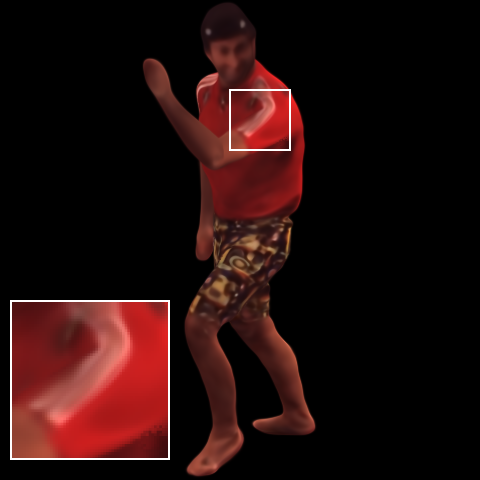}%
}\\%
\fbox{\includegraphics%
[width=\hmqualscale]
{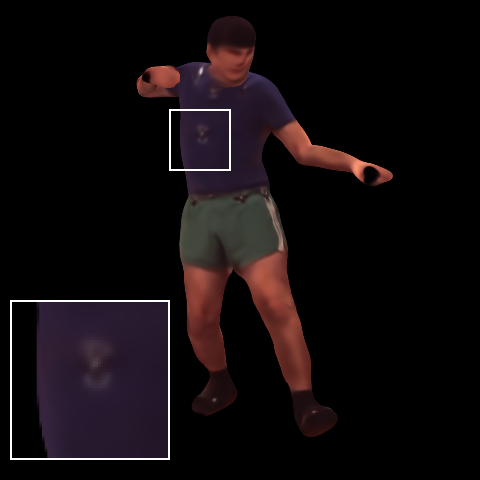}%
}\\%
{ \hmqualmb}%
}%
\hfill
\parbox[t]{\hmqualscale}{%
\centering%
\fbox{\includegraphics%
[width=\hmqualscale]
{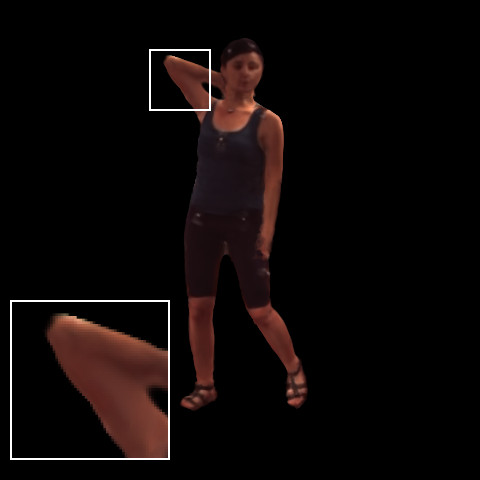}%
}\\%
\fbox{\includegraphics%
[width=\hmqualscale]
{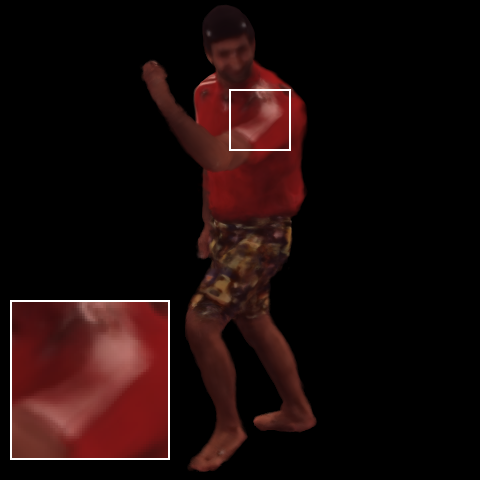}%
}\\%
\fbox{\includegraphics%
[width=\hmqualscale]
{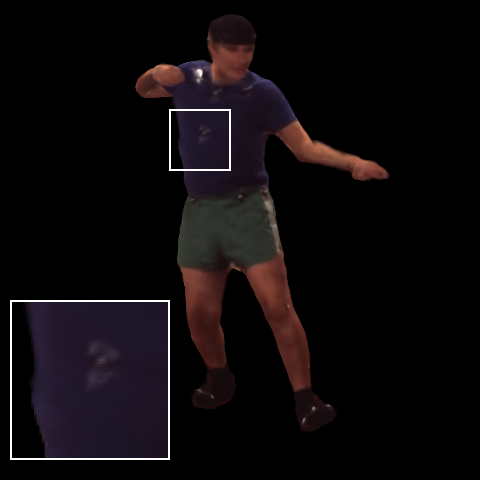}%
}\\%
{ \hmqualmc}%
}%
\hfill
\parbox[t]{\hmqualscale}{%
\centering%
\fbox{\includegraphics%
[width=\hmqualscale]
{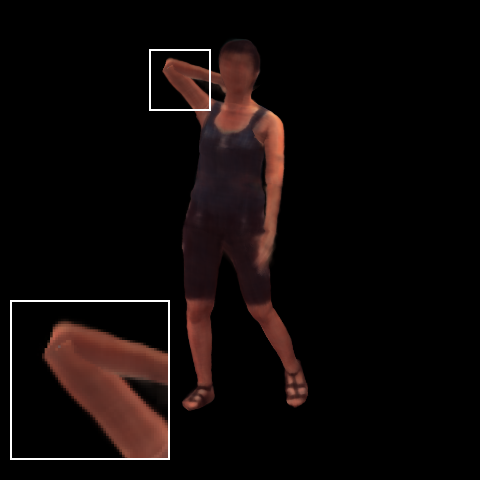}%
}\\%
\fbox{\includegraphics%
[width=\hmqualscale]
{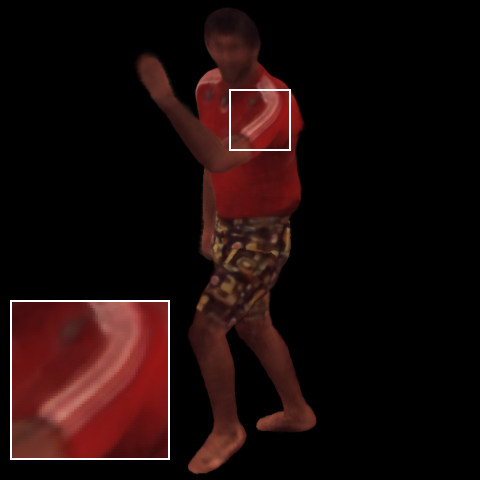}%
}\\%
\fbox{\includegraphics%
[width=\hmqualscale]
{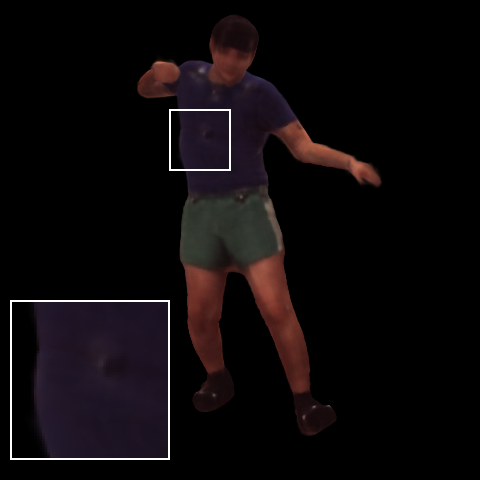}%
}\\%
{ \hmqualmd}%
}%
\hfill%
\parbox[t]{\hmqualscale}{%
\centering%
\fbox{\includegraphics%
[width=\hmqualscale]
{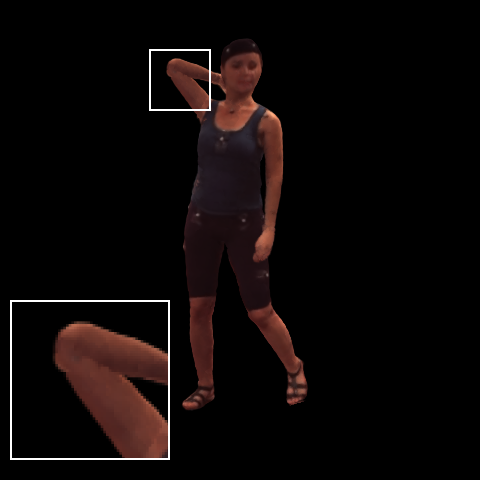}%
}\\%
\fbox{\includegraphics%
[width=\hmqualscale]
{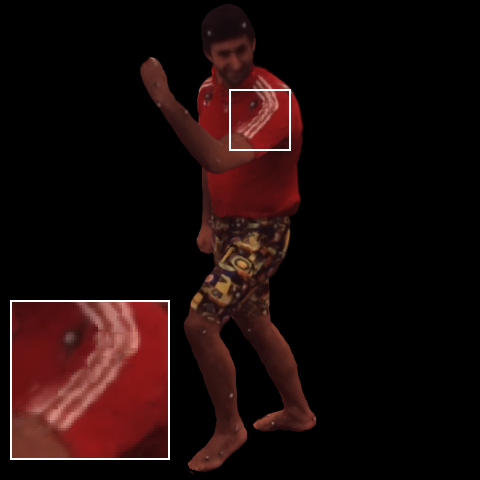}%
}\\%
\fbox{\includegraphics%
[width=\hmqualscale]
{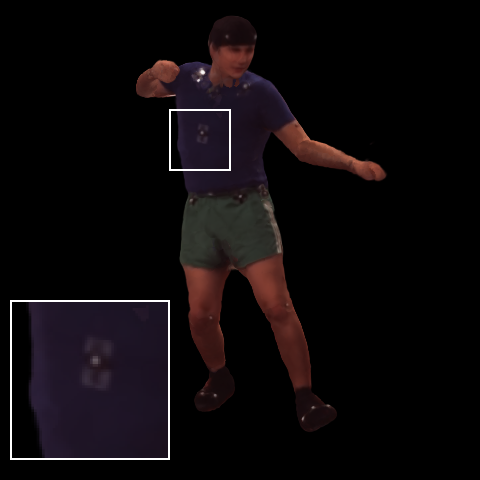}%
}\\%
{\ourapproach~(Ours)}
}%
\hfill
\parbox[t]{\hmqualscale}{%
\centering%
\fbox{\includegraphics%
[width=\hmqualscale]
{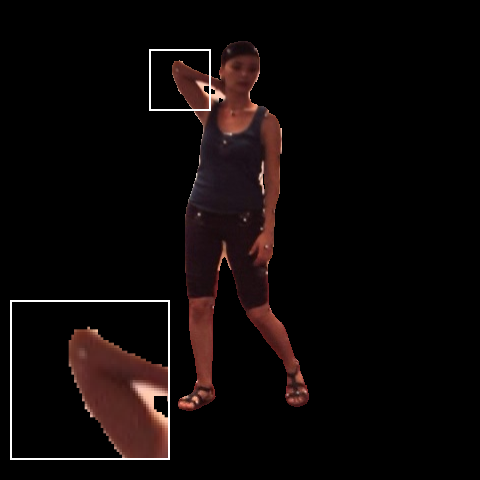}%
}\\%
\fbox{\includegraphics%
[width=\hmqualscale]
{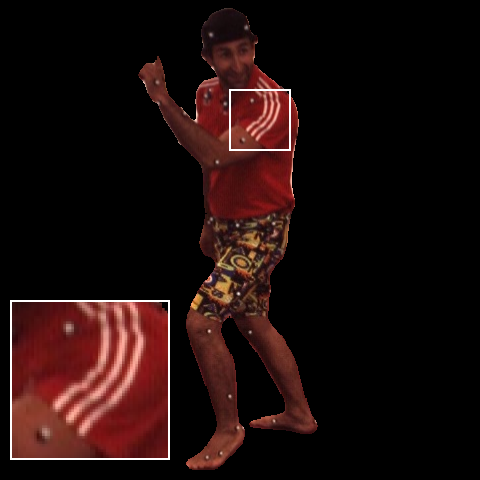}%
}\\%
\fbox{\includegraphics%
[width=\hmqualscale]
{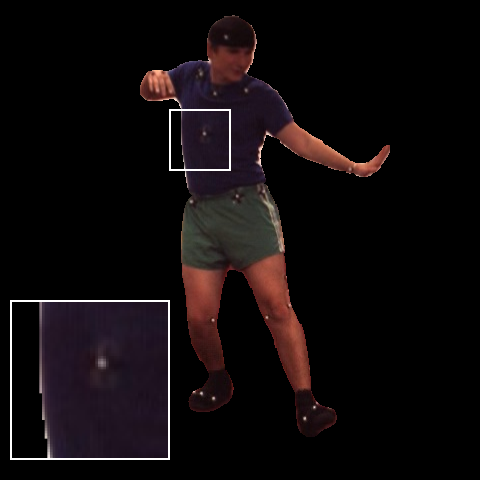}%
}\\%
{ Ground truth}%
}%
\centering%
\caption{\textbf{Unseen pose synthesis on Anim-NeRF Human3.6M~\cite{Ionescu11,Ionescu14a,peng2021animatable} test split}. Our ~\ourapproach~shows better appearance consistency on unseen poses, with improved articulation, sharper textures, and fine details like motion trackers and stripes.
}%
\label{fig:exp-qualitative-novel-pose}
\end{figure*}
\section{Experiments}
\begin{table*}[t]
\caption{\textbf{Novel-view synthesis comparisons on Anim-NeRF Human3.6M~\cite{Ionescu11,Ionescu14a,peng2021animatable} test split.} Our NPC benefits from the point-based representation, and achieves better overall perceptual quality.}
\centering
\resizebox{\linewidth}{!}{
\setlength{\tabcolsep}{0pt}
\begin{tabular}{lcccccccccccccccccccccccc}
\toprule
& \multicolumn{3}{c}{S1} & \multicolumn{3}{c}{S5} & \multicolumn{3}{c}{S6} & \multicolumn{3}{c}{S7} & \multicolumn{3}{c}{S8} & \multicolumn{3}{c}{S9} & \multicolumn{3}{c}{S11} & \multicolumn{3}{c}{Avg} \\
\cmidrule(lr){2-4}\cmidrule(lr){5-7}\cmidrule(lr){8-10}\cmidrule(lr){11-13}\cmidrule(lr){14-16}\cmidrule(lr){17-19}\cmidrule(lr){20-22}\cmidrule(lr){23-25}%
  & PSNR$\uparrow$  & SSIM$\uparrow$  & LPIPS$\downarrow$  & ~PSNR\phantom{$\uparrow$}& SSIM\phantom{$\uparrow$}  & LPIPS\phantom{$\uparrow$}  & ~PSNR\phantom{$\uparrow$}& SSIM\phantom{$\uparrow$}  & LPIPS\phantom{$\uparrow$}  & ~PSNR\phantom{$\uparrow$}& SSIM\phantom{$\uparrow$}  & LPIPS\phantom{$\uparrow$}  & PSNR$\uparrow$  & SSIM$\uparrow$  & LPIPS$\downarrow$  & ~PSNR\phantom{$\uparrow$}& SSIM\phantom{$\uparrow$}  & LPIPS\phantom{$\uparrow$}  & ~PSNR\phantom{$\uparrow$}& SSIM\phantom{$\uparrow$}  & LPIPS\phantom{$\uparrow$}  & ~PSNR\phantom{$\uparrow$}& SSIM\phantom{$\uparrow$}  & LPIPS\phantom{$\uparrow$} \\
\multicolumn{25}{l}{\textbf{Template/Scan-based prior}}\\
NeuralBody& 22.88& 0.897& 0.139& 24.61& 0.917& 0.128& 22.83& 0.888& 0.155& 23.17& 0.915& 0.132& 21.72& 0.894& 0.151& 24.29& 0.911& 0.122& 23.70& 0.896& 0.168& 23.36& 0.905& 0.140\\
\rowcolor{Gray}
Anim-NeRF& 22.74& 0.896& 0.151& 23.40& 0.895& 0.159& 22.85& 0.871& 0.187& 21.97& 0.891& 0.161& 22.82& 0.900& 0.146& 24.86& 0.911& 0.145& 24.76& 0.907& 0.161& 23.34& 0.897& 0.157\\
ARAH$^\dagger$& 24.53& 0.921& 0.103& 24.67& 0.921& 0.115& 24.37& 0.904& 0.133& 24.41& 0.922& 0.115& \textbf{24.15}& \textbf{0.924}& \textbf{0.104}& 25.43& 0.924& 0.112& 24.76& 0.918& 0.128& 24.63& 0.920& 0.115\\
\midrule
\multicolumn{25}{l}{\textbf{Template-free}}\\
A-NeRF& 23.93& 0.912& 0.118& 24.67& 0.919& 0.114& 23.78& 0.887& 0.147& 24.40& 0.917& 0.125& 22.70& 0.907& 0.130& 25.58& 0.916& 0.126& 24.38& 0.905& 0.152& 24.26& 0.911& 0.129\\
\rowcolor{Gray}
DANBO& 23.95& 0.916& 0.108& 24.86& 0.924& 0.108& 24.54& 0.903& 0.129& 24.45& 0.920& 0.113& 23.36& 0.917& 0.116& 26.15& 0.925& 0.108& 25.58& 0.917& 0.127& 24.73& 0.918& 0.115\\
TAVA& \textbf{25.28}& \textbf{0.928}& 0.108& 24.00& 0.916& 0.122& 23.44& 0.894& 0.138& 24.25& 0.916& 0.130& 23.71& 0.921& 0.116& 26.20& 0.923& 0.119& \textbf{26.17}& \textbf{0.928}& 0.133& 24.72& 0.919& 0.124\\
\midrule
\rowcolor{Gray}
\textbf{\ourapproach~(Ours)}& 24.81& 0.922& \textbf{0.097}& \textbf{24.92}& \textbf{0.926}& \textbf{0.100}& \textbf{24.89}& \textbf{0.909}& \textbf{0.118}& \textbf{24.87}& \textbf{0.924}& \textbf{0.105}& 24.03& 0.923& 0.104& \textbf{26.39}& \textbf{0.930}& \textbf{0.095}& 25.86& 0.925& \textbf{0.117}& \textbf{25.13}& \textbf{0.924}& \textbf{0.104}\\
\multicolumn{25}{l}{$^\dagger$: we evaluate using the officially released ARAH, which has undergone refactorization, resulting in slightly different numbers to the ones in~\cite{wang2022arah}.}\\
\end{tabular}
\label{tab:exp-h36m-novel-view}
}
\end{table*}

\begin{table*}[t]
\caption{\textbf{Unseen pose rendering comparisons on Anim-NeRF Human3.6M~\cite{Ionescu11,Ionescu14a,peng2021animatable} test split.}The efficient canonical feature mapping carried out by our point-based approach enables better generalization in unseen poses.}
\centering
\resizebox{\linewidth}{!}{
\setlength{\tabcolsep}{0pt}
\begin{tabular}{lcccccccccccccccccccccccc}
\toprule
& \multicolumn{3}{c}{S1} & \multicolumn{3}{c}{S5} & \multicolumn{3}{c}{S6} & \multicolumn{3}{c}{S7} & \multicolumn{3}{c}{S8} & \multicolumn{3}{c}{S9} & \multicolumn{3}{c}{S11} & \multicolumn{3}{c}{Avg} \\
\cmidrule(lr){2-4}\cmidrule(lr){5-7}\cmidrule(lr){8-10}\cmidrule(lr){11-13}\cmidrule(lr){14-16}\cmidrule(lr){17-19}\cmidrule(lr){20-22}\cmidrule(lr){23-25}%
  & PSNR$\uparrow$  & SSIM$\uparrow$  & LPIPS$\downarrow$  & ~PSNR\phantom{$\uparrow$}& SSIM\phantom{$\uparrow$}  & LPIPS\phantom{$\uparrow$}  & ~PSNR\phantom{$\uparrow$}& SSIM\phantom{$\uparrow$}  & LPIPS\phantom{$\uparrow$}  & ~PSNR\phantom{$\uparrow$}& SSIM\phantom{$\uparrow$}  & LPIPS\phantom{$\uparrow$}  & PSNR$\uparrow$  & SSIM$\uparrow$  & LPIPS$\downarrow$  & ~PSNR\phantom{$\uparrow$}& SSIM\phantom{$\uparrow$}  & LPIPS\phantom{$\uparrow$}  & ~PSNR\phantom{$\uparrow$}& SSIM\phantom{$\uparrow$}  & LPIPS\phantom{$\uparrow$}  & ~PSNR\phantom{$\uparrow$}& SSIM\phantom{$\uparrow$}  & LPIPS\phantom{$\uparrow$} \\
\multicolumn{25}{l}{\textbf{Template/Scan-based prior}}\\
NeuralBody& 22.10& 0.878& 0.143& 23.52& 0.897& 0.144& 23.42& 0.892& 0.146& 22.59& 0.893& 0.163& 20.94& 0.876& 0.172& 23.05& 0.885& 0.150& 23.72& 0.884& 0.179& 22.81& 0.888& 0.157\\
\rowcolor{Gray}
Anim-NeRF& 21.37& 0.868& 0.167& 22.29& 0.875& 0.171& 22.59& 0.884& 0.159& 22.22& 0.878& 0.183& 21.78& 0.882& 0.162& 23.73& 0.886& 0.157& 23.92& 0.889& 0.176& 22.61& 0.881& 0.170\\
ARAH$^\dagger$& 23.18& 0.903& 0.116& 22.91& 0.894& 0.133& 23.91& 0.901& 0.125& 22.72& 0.896& 0.143& 22.50& 0.899& 0.128& 24.15& 0.896& 0.135& 23.93& 0.899& 0.143& 23.27& 0.897& 0.134\\
\midrule
\multicolumn{25}{l}{\textbf{Template-free}}\\
A-NeRF& 22.67& 0.883& 0.159& 22.96& 0.888& 0.155& 22.77& 0.869& 0.170& 22.80& 0.880& 0.182& 21.95& 0.886& 0.170& 24.16& 0.889& 0.164& 23.40& 0.880& 0.190& 23.02& 0.883& 0.171\\
\rowcolor{Gray}
DANBO& 23.03& 0.895& 0.121& \textbf{23.66}& 0.903& 0.124& 24.57& 0.906& 0.118& 23.08& 0.897& 0.139& 22.60& 0.904& 0.132& 24.79& 0.904& 0.130& 24.57& 0.901& 0.146& 23.74& 0.901& 0.131\\
TAVA& \textbf{23.83}& \textbf{0.908}& 0.120& 22.89& 0.898& 0.135& 24.54& 0.906& 0.122& 22.33& 0.882& 0.163& 22.50& 0.906& 0.130& 24.80& 0.901& 0.138& \textbf{25.22}& \textbf{0.913}& 0.145& 23.52& 0.899& 0.141\\
\midrule
\rowcolor{Gray}
\textbf{\ourapproach~(Ours)} & 23.39& 0.901& \textbf{0.109}& 23.63& \textbf{0.906}& \textbf{0.113}& \textbf{24.59}& \textbf{0.911}& \textbf{0.105}& \textbf{23.46}& \textbf{0.903}& \textbf{0.129}& \textbf{22.87}& \textbf{0.907}& \textbf{0.121}& \textbf{24.86}& \textbf{0.907}& \textbf{0.115}& 25.13& 0.911& \textbf{0.130}& \textbf{23.96}& \textbf{0.906}& \textbf{0.119}\\
\multicolumn{25}{l}{$^\dagger$: we evaluate using the officially released ARAH, which has undergone refactorization, resulting in slightly different numbers to the ones in~\cite{wang2022arah}. 
}\\
\end{tabular}
\label{tab:exp-h36m-novel-pose}
}
\end{table*}

\begin{table}[t]
\caption{\textbf{Unseen pose synthesis on MonoPerfCap~\cite{Xu18a}}. ~\ourapproach~ shows better overall perceptual quality over DANBO on learning generalized model from monocular videos. 
}
\centering
\resizebox{1.0\linewidth}{!}{
\setlength{\tabcolsep}{3pt}
\begin{tabular}{lcccccc}
\toprule
& \multicolumn{2}{c}{ND} & \multicolumn{2}{c}{WP} & \multicolumn{2}{c}{Avg} \\
\cmidrule(lr){2-3}\cmidrule(lr){4-5}\cmidrule(lr){6-7}%
 & KIDx100~$\downarrow$  & LPIPS~$\downarrow$  & KIDx100~$\downarrow$  & LPIPS~$\downarrow$  & KIDx100~$\downarrow$  & LPIPS~$\downarrow$ \\
\midrule
A-NeRF& 4.97& 0.197& 6.53& 0.223& 5.75& 0.208\\
\rowcolor{Gray}
DANBO& 4.83& \textbf{0.194}& 4.66& 0.214& 4.74& 0.202\\
\midrule
\textbf{\ourapproach~(Ours)}& \textbf{2.57}& 0.198& \textbf{3.56}& \textbf{0.207}& \textbf{3.07}& \textbf{0.202}\\
\end{tabular}
\label{tab:exp-perfcap-novel-pose}
}
\end{table}

\newlength\expretargetscale
\setlength\expretargetscale{0.20\linewidth}
\ifblurface
\newcommand{\expretargetpost}{_blur}
\else
\newcommand{\expretargetpost}{}
\fi
\newcommand{\expretargetpath}{statics/figs/exp/retarget/}
\newcommand{\expretargetma}{DANBO}
\newcommand{\expretargetmb}{Ours}
\newcommand{\expretargeta}{nadia}
\newcommand{\expretargetanum}{01}
\newcommand{\expretargetb}{nadia}
\newcommand{\expretargetbnum}{00}
\newcommand{\expretargetc}{S7}
\newcommand{\expretargetcnum}{00}
\newcommand{\expretargetd}{weipeng}
\newcommand{\expretargetdnum}{00}
\newcommand{\expretargete}{S1}
\newcommand{\expretargetenum}{00}
\begin{figure}[t]
\setlength{\fboxsep}{0pt}%
\setlength{\fboxrule}{0pt}%
\parbox[t]{\expretargetscale}{%
\centering%
\fbox{\includegraphics%
[width=\expretargetscale,trim=105 0 107 0,clip]%
{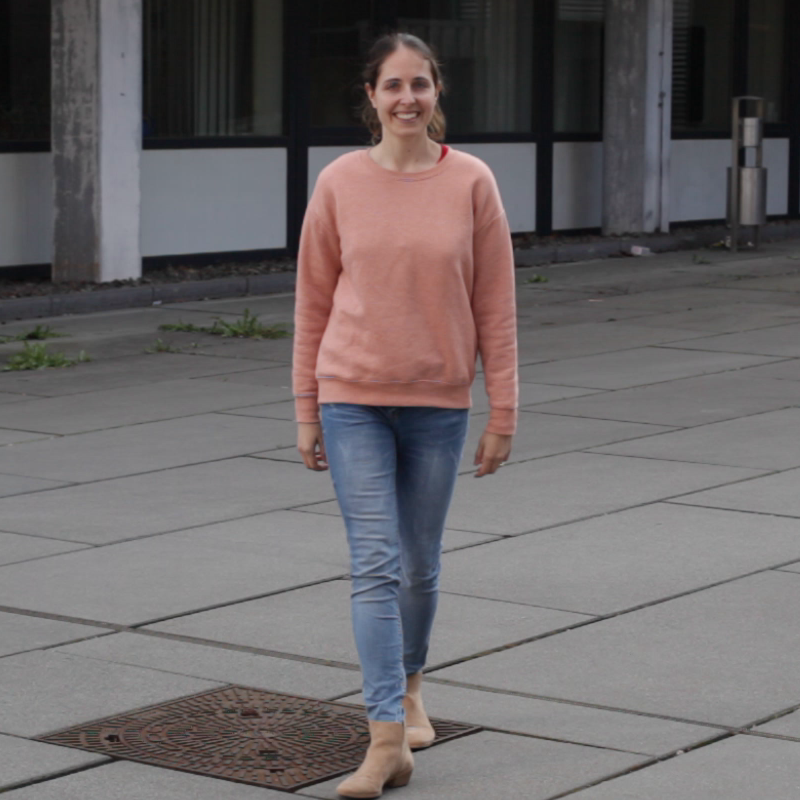}%
}\\%
\fbox{\includegraphics%
[width=\expretargetscale,trim=120 0 120 0,clip]%
{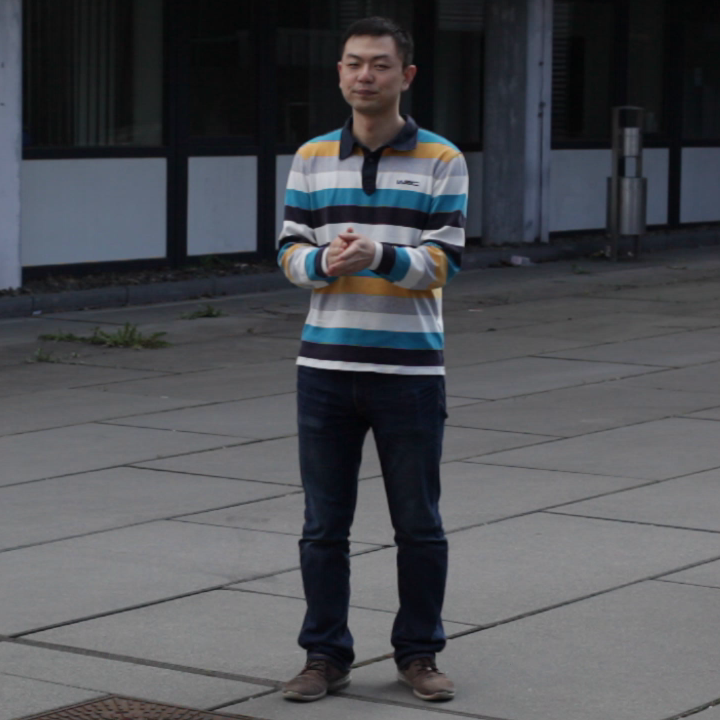}%
}\\%
{\scriptsize Training example}%
}%
\hfill
\parbox[t]{\expretargetscale}{%
\centering%
\fbox{\includegraphics%
[width=\expretargetscale,trim=80 0 80 0,clip]%
{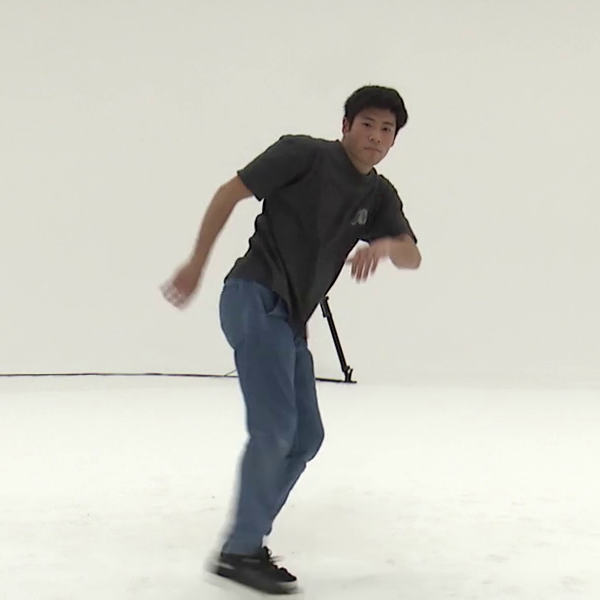}%
}\\%
\fbox{\includegraphics%
[width=\expretargetscale,trim=80 0 80 0,clip]%
{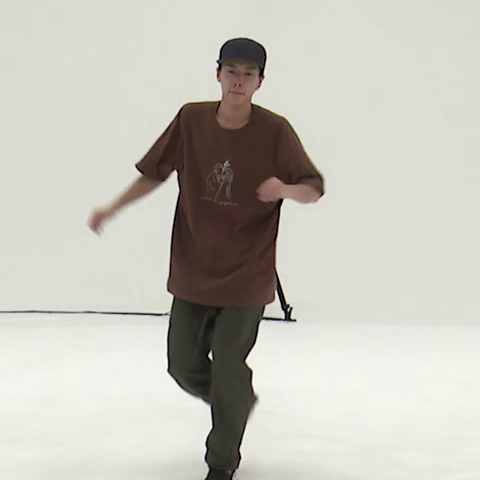}%
}\\%
{\footnotesize Reference pose}%
}%
\hfill%
\parbox[t]{\expretargetscale}{%
\centering%
\fbox{\includegraphics%
[width=\expretargetscale,trim=80 0 80 0,clip]%
{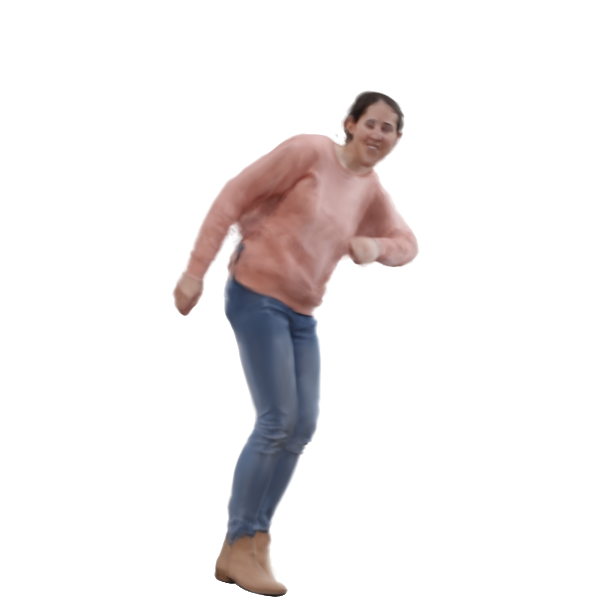}%
}\\%
\fbox{\includegraphics%
[width=\expretargetscale,trim=80 0 80 0,clip]%
{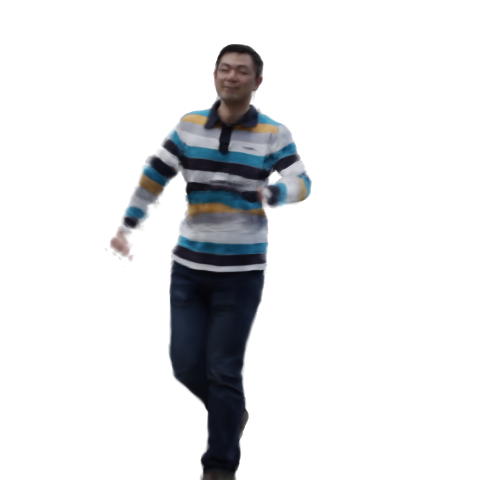}%
}\\%
{\footnotesize \expretargetma}%
}%
\hfill%
\parbox[t]{\expretargetscale}{%
\centering%
\fbox{\includegraphics%
[width=\expretargetscale,trim=80 0 80 0,clip]%
{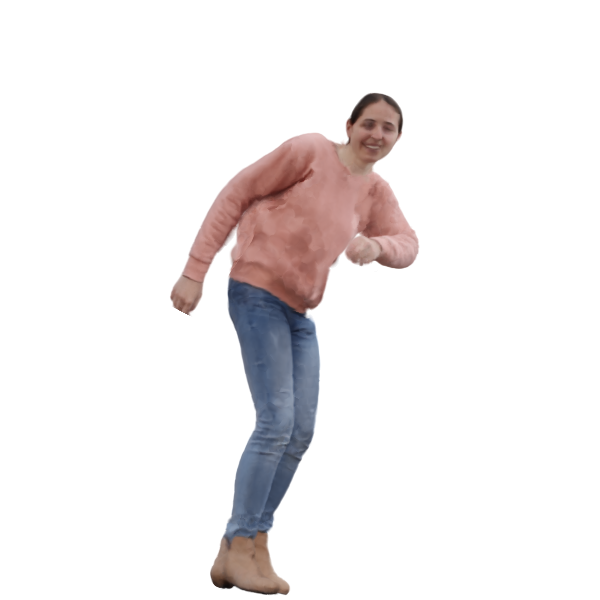}%
}\\%
\fbox{\includegraphics%
[width=\expretargetscale,trim=80 0 80 0,clip]%
{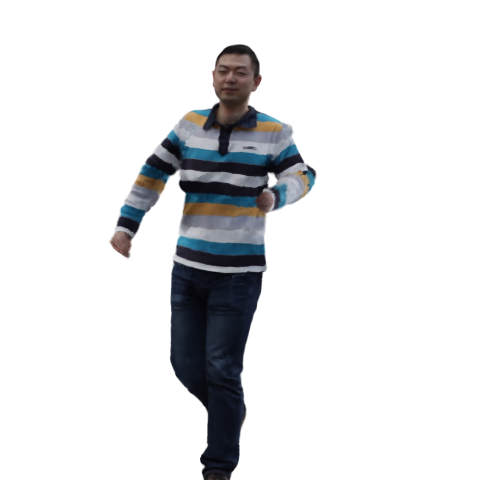}%
}\\%
{\footnotesize NPC (Ours)}%
}%
\centering%
\caption{\textbf{Motion retargeting from out-of-distribution poses on various subjects}. \ourapproach~retains better appearance consistency and texture detail.
}%
\label{fig:exp-retarget}
\end{figure}

We quantify the improvements our~\ourapproach{} brings over the most recent surface-free approach TAVA~\cite{li2022tava}, DANBO~\cite{su2022danbo} and A-NeRF~\cite{su2021anerf}, as well as most recent and established approaches that leverage template or scan-based prior, including ARAH~\cite{wang2022arah}, Anim-NeRF~\cite{peng2021animatable} and NeuralBody~\cite{peng2020neuralbody}. Moreover, we conduct ablation studies to verify how our proposed per-point encoding, pose-dependent feature, and coordinate features help improve perceptual quality. The supplemental video and document provide additional qualitative results and implementation details. The code will be made available upon publication to facilitate human body modeling research\footnote{All data sourcing, modeling codes, and experiments were developed at University of British Columbia. Meta did not obtain the data/codes or conduct any experiments in this work.}.

\parag{Metrics.} %
We adopt standard image metrics, including pixel-wise PSNR and SSIM~\cite{wang2004ssim}, as well as perceptual metrics LPIPS~\cite{zhang2018perceptual} and KID~\cite{binkowski2018demystifying,parmar2021cleanfid} that better quantify the improvement on texture details under slight misalignment that is unavoidable when driving detailed characters with only sparse pose. %
All scores are on the holdout testing set. Specifically, we evaluate the performance for \textit{ novel-view synthesis} on multi-view datasets by rendering seen poses from unseen testing views and \textit{novel-pose synthesis} to synthesize appearances of body poses unseen during training.

\parag{Datasets.} We evaluate ~\ourapproach~on the established benchmarks for body modeling, including both indoor, outdoor, monocular video capture, and synthetic animal datasets.
\begin{itemize}[noitemsep,nolistsep]
    \item Human3.6M~\cite{Ionescu11,Ionescu14a}: we use the protocol from Anim-NeRF~\cite{peng2021animatable}, evaluating on a total of 7 subjects using the foreground maps provided by the authors.
    \item MonoPerfCap~\cite{Xu18a}: the dataset consists of multiple monocular outdoor footages. We use the same 2 subjects as in~\cite{su2021anerf}. 
    The training poses and cameras are SPIN estimates~\cite{kolotouros2019learning_spin} refined by A-NeRF.
\end{itemize}

We additionally evaluate \ourapproach{} on one subject with loose clothing from ZJU-Mocap~\cite{fang2021mirrored,peng2020neuralbody}, and use challenging motion sequences, including dancing and gymnastic from poses AIST++~\cite{li2021learn} and SURREAL+CMU-Mocap~\cite{CMUMOCAP,varol17_surreal} dataset, for animating our learned characters. Finally, we use the synthetic wolf dataset from TAVA~\cite{li2022tava} as a proof-of-concept on how we can apply~\ourapproach~to non-human subjects similar to other template-free methods~\cite{li2022tava,noguchi2021narf,su2022danbo}. 

\subsection{Novel View Synthesis}
Our point-based representation anchors high-frequency local details that are shared across all training views, enabling improved visual details even when rendering from a novel view not presented in the training data.
Compared to ARAH~\cite{wang2022arah} using implicit-surface prior~\cite{wang2021metaavatar} and root-finding for canonical mapping, ~\ourapproach~ synthesizes sharper results despite using only noisy surface points extracted without supervision.~\figref{fig:exp-qualitative-novel-view} shows that we recover better appearance details such as markers and shoelaces. We observe that TAVA, although using root-finding like ARAH for mapping, suffers more blurry artifacts. We conjecture that the root-finding algorithm requires good initialization for video sequences with complex motion.~\tabref{tab:exp-h36m-novel-view} quantifies the improvement over the recent template-free and template/scan-based neural body fields.

\subsection{Unseen Pose and Animating NPC}
Unseen pose synthesis requires the method to maintain consistent appearances in out-of-distribution pose configurations. Our NPC shows overall more fine-grained and consistent renderings as shown in ~\figref{fig:exp-qualitative-novel-pose}. Compared to Neural Body~\cite{peng2020neuralbody}, which anchors the body representation on template body mesh and diffuses the feature using 3D CNN, our NPC produces crispier details like motion capture body trackers and wrinkles, and maintains better texture consistency like the clearer stripes. We attribute this to our point-based feature encoding and explicit non-deformation on surface points that, in combination, enable better localization of the body features. \tabref{tab:exp-h36m-novel-pose} and \tabref{tab:exp-perfcap-novel-pose} verify the consistent improvement of our ~\ourapproach~ on the perceptual quality on not just multi-view dataset Human3.6M~\cite{Ionescu11,Ionescu14a}, but also outdoor monocular video sequences MonoPerfCap~\cite{Xu18a}. Note that we omit PSNR and SSIM for~\tabref{tab:exp-perfcap-novel-pose} as these metrics are susceptible to the varying lighting conditions in outdoor sequences.

We further animate the learned NPC models with motion sequences from out-of-distribution body poses in ~\figref{fig:exp-retarget}, to showcase how we can potentially apply our point-based characters for animation. Note that we cannot quantify the performance for these examples as there are no ground truth images available for evaluation.

\parag{Geometry Comparisons.}
\ourapproach~reconstructs detailed 3D geometry even from monocular video and generalizes deformations well to unseen poses, with an overall more complete body outline, as visualized in~\figref{fig:exp-geometry}.  Different to DANBO and other implicit methods, our point scaffold provides correspondences across frames and is therefore applicable to dense 3D pose estimation. Finally, we show that \ourapproach{} can recover imperfect geometry, such as correcting the missing foot and trouser shape in \figref{fig:supp-pts-correct}.

\subsection{\ourapproach{} Deformation on Loose Clothing}
\newlength\expgeometryscale
\setlength\expgeometryscale{0.160\linewidth}
\ifblurface%
\newcommand{\expgeometrypost}{_blur}
\else%
\newcommand{\expgeometrypost}{}
\fi%
\begin{figure}[t]
\setlength{\fboxsep}{0pt}%
\setlength{\fboxrule}{0pt}%
\parbox[t]{\expgeometryscale}{%
\centering%
\fbox{\includegraphics%
[width=\expgeometryscale,trim=400 300 1150 100,clip]%
{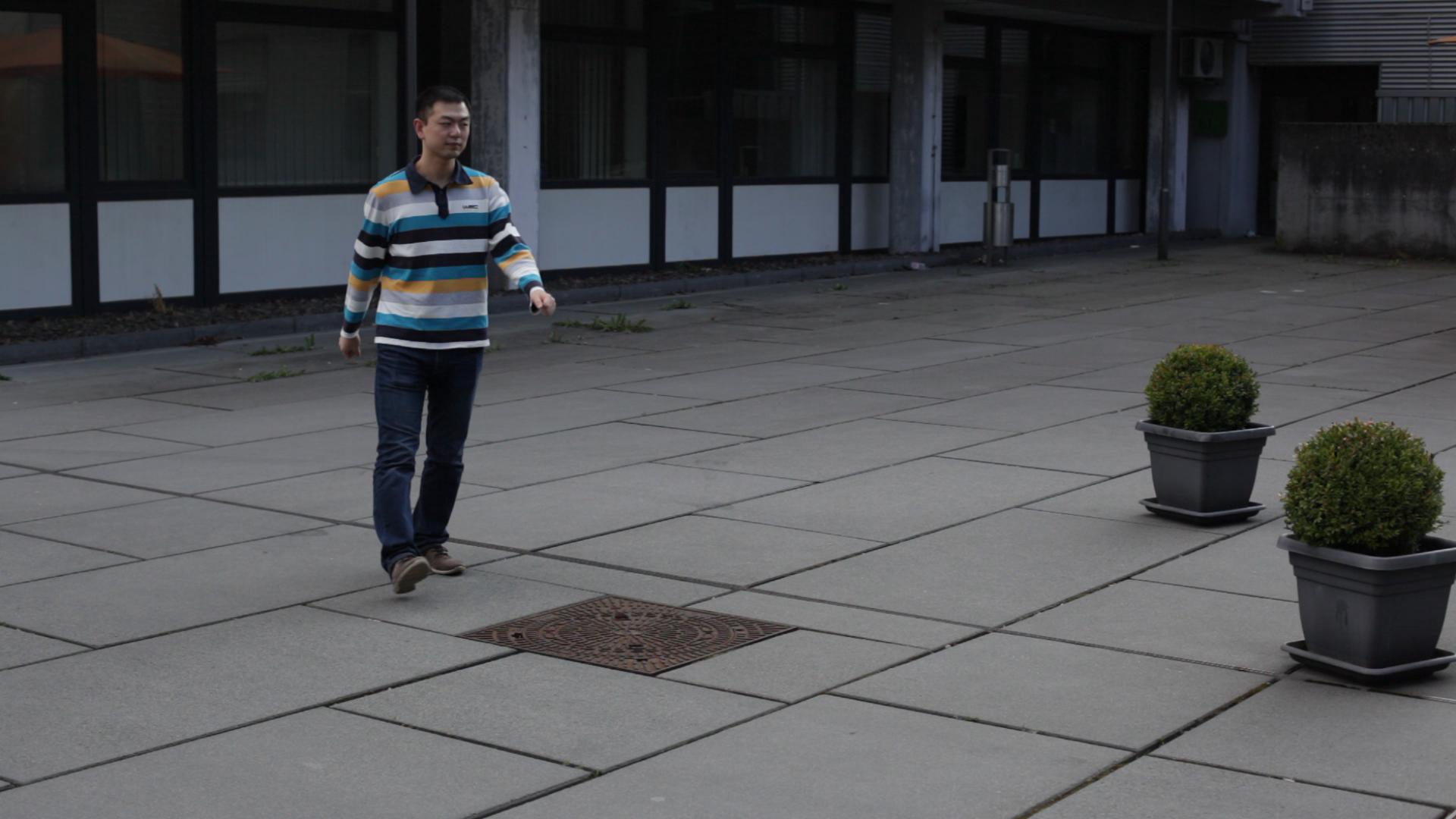}%
}\\%
{\footnotesize  GT}%
}%
\hfill%
\parbox[t]{\expgeometryscale}{%
\centering%
\fbox{\includegraphics%
[width=\expgeometryscale,trim=310 50 230 50,clip]%
{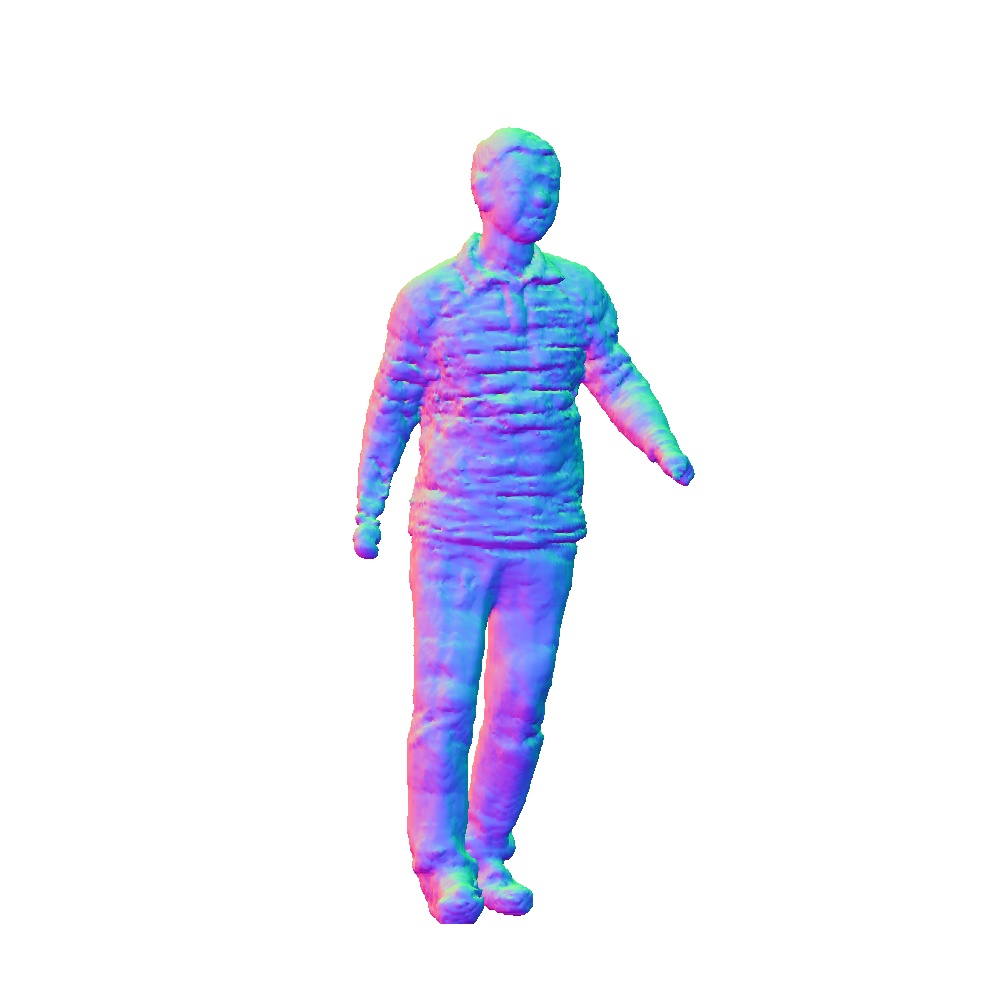}%
}\\
{\footnotesize DANBO}%
}%
\hfill%
\parbox[t]{\expgeometryscale}{%
\centering%
\fbox{\includegraphics%
[width=\expgeometryscale,trim=310 50 230 50,clip]%
{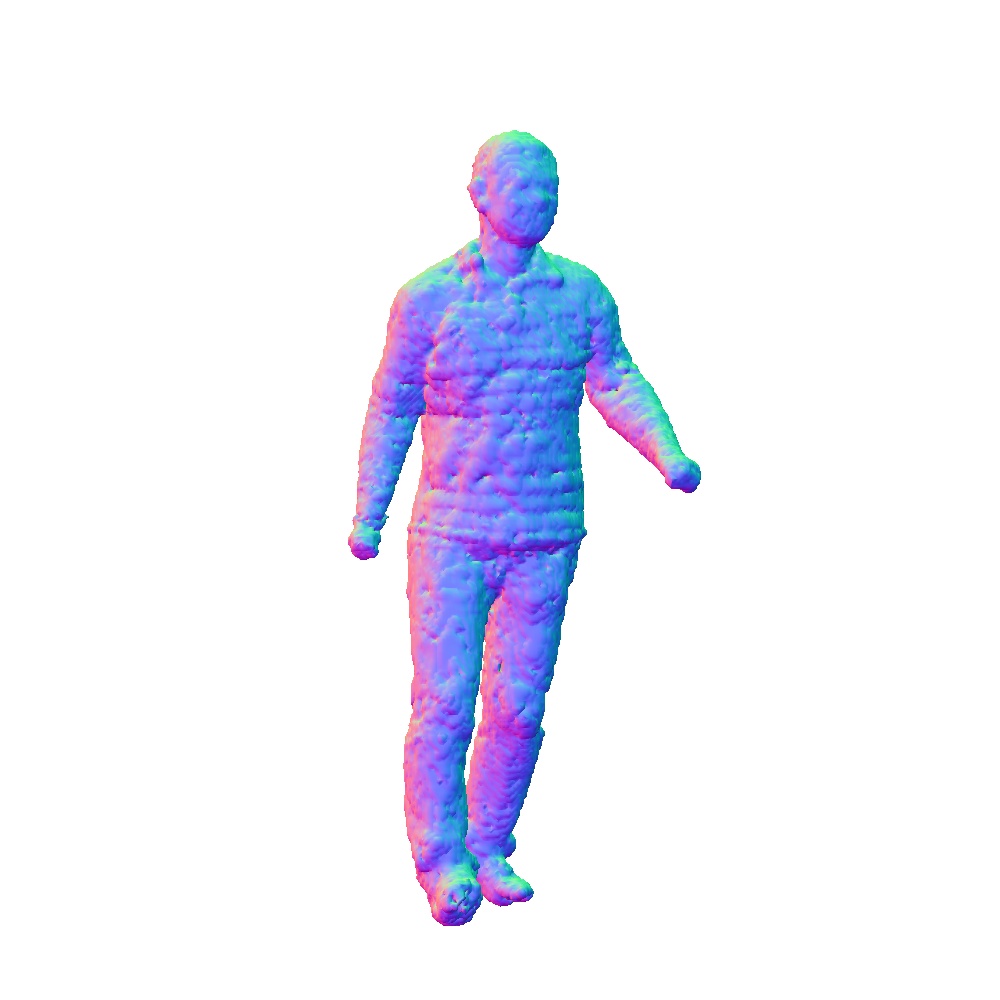}%
}\\%
{\footnotesize \ourapproach (Ours)}%
}%
\hfill%
\parbox[t]{\expgeometryscale}{%
\centering%
\fbox{\includegraphics%
[width=\expgeometryscale,trim=120 0 130 0,clip]%
{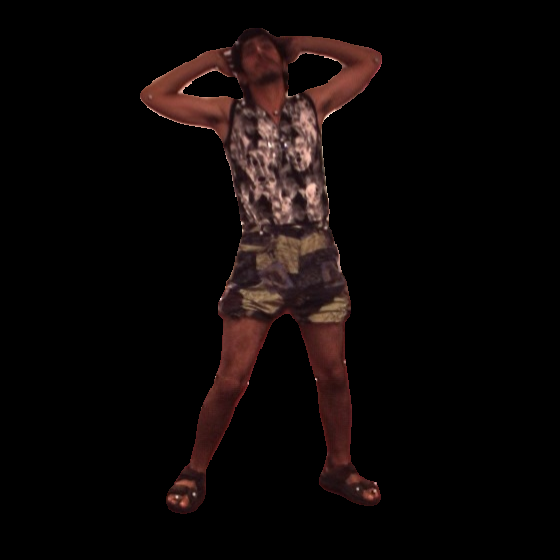}%
}\\%
{\footnotesize GT}%
}%
\hfill%
\parbox[t]{\expgeometryscale}{%
\centering%
\fbox{\includegraphics%
[width=\expgeometryscale,trim=240 20 270 100,clip]%
{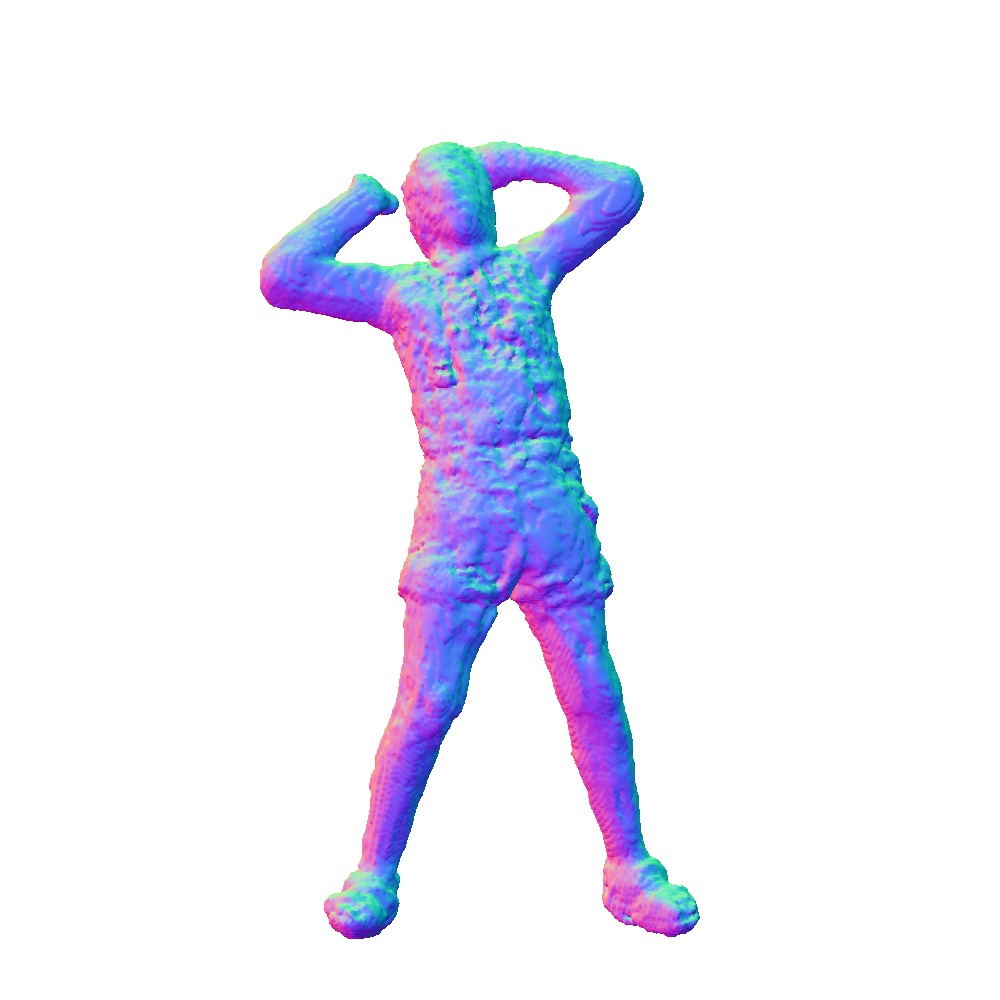}%
}\\%
{\footnotesize DANBO}%
}%
\hfill%
\parbox[t]{\expgeometryscale}{%
\centering%
\fbox{\includegraphics%
[width=\expgeometryscale,trim=240 20 270 100,clip]%
{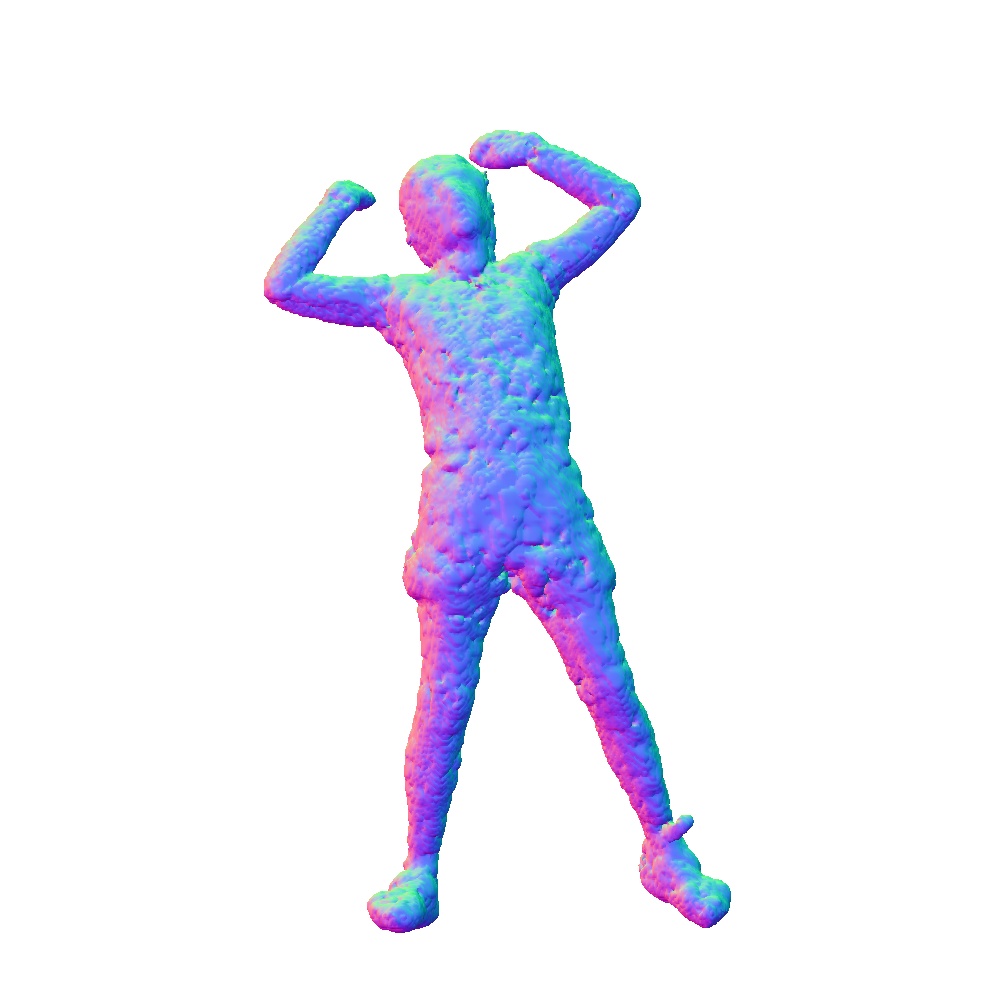}%
}\\%
{\footnotesize \ourapproach~(Ours)}%
}%
\centering%
\caption{\textbf{Body geometry in unseen poses.} ~\ourapproach~matches the shape quality of DANBO. %
}%
\label{fig:exp-geometry}
\end{figure}
\newlength\expptscorrectscale
\setlength\expptscorrectscale{0.28\linewidth}
\ifblurface
\newcommand{\expptscorrectpost}{_blur}
\else
\newcommand{\expptscorrectpost}{}
\fi
\newcommand{\expptscorrectpath}{statics/figs/supp/point_clouds_correct/}
\begin{figure}[t]
\setlength{\fboxsep}{0pt}%
\setlength{\fboxrule}{0pt}%
\parbox[t]{\expptscorrectscale}{%
\centering%
\fbox{\includegraphics%
[width=\expptscorrectscale,trim=10 0 16 0,clip]%
{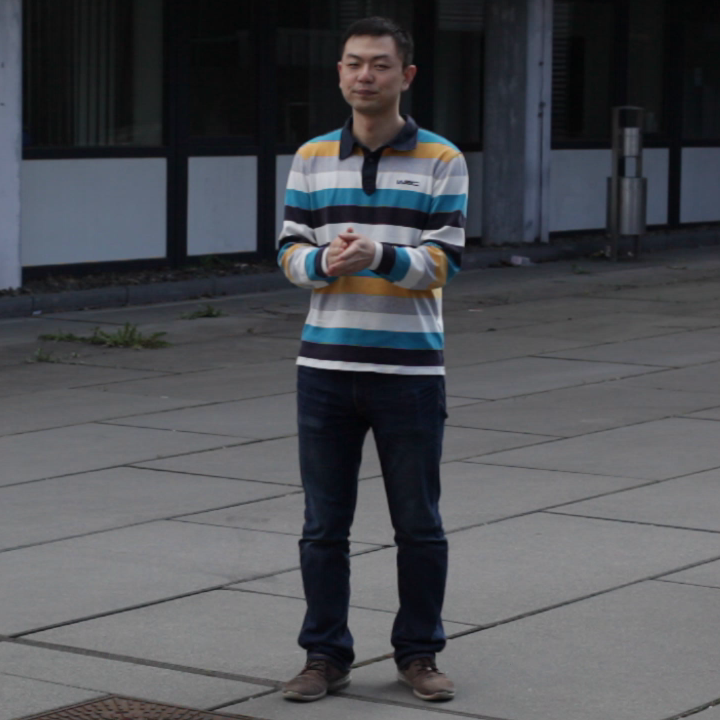}%
}\\%
{\scriptsize Reference}%
}%
\hfill
\parbox[t]{\expptscorrectscale}{%
\centering%
\fbox{\includegraphics%
[width=\expptscorrectscale,trim=360 250 200 300,clip]%
{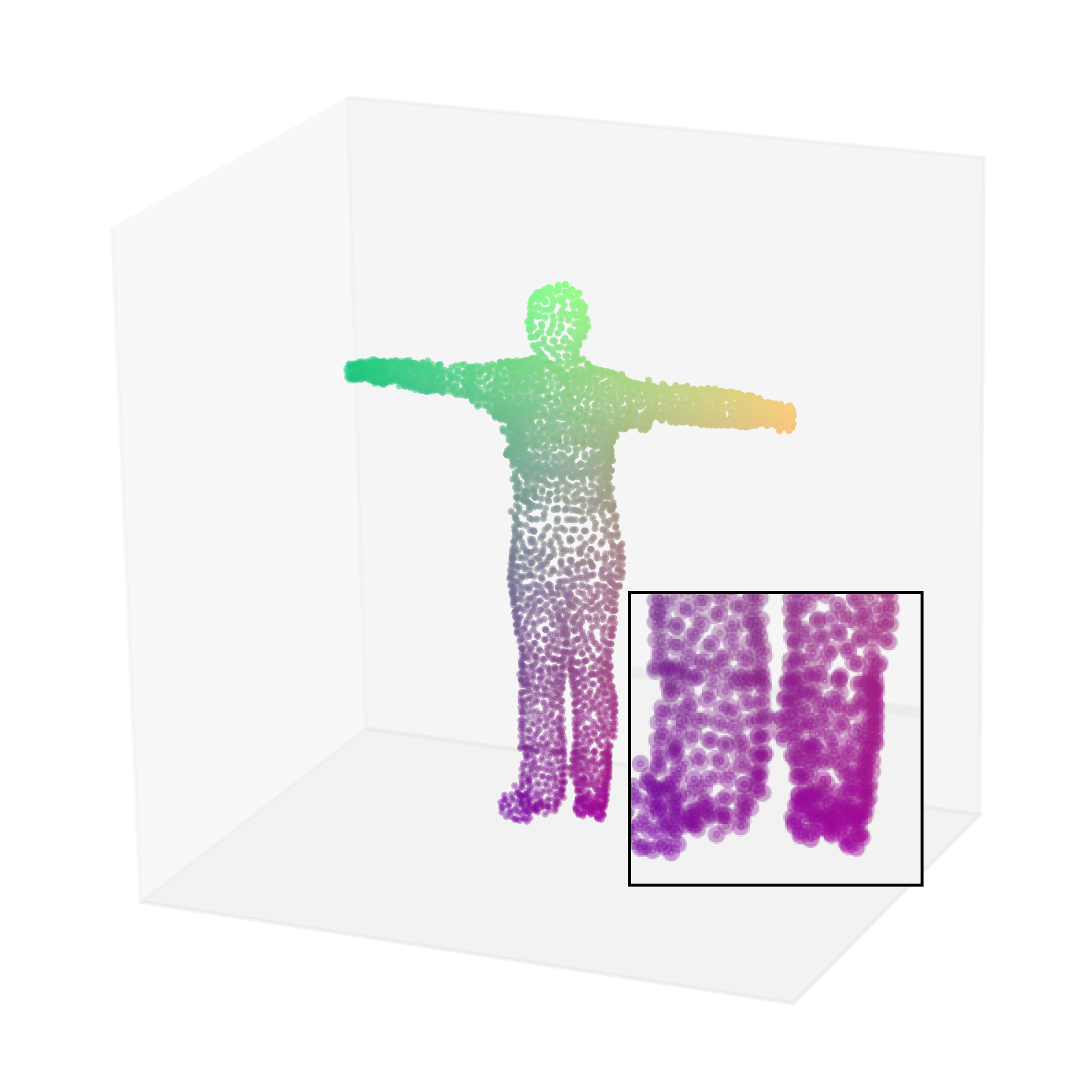}%
}\\%
{\footnotesize Initial points}%
}%
\hfill%
\parbox[t]{\expptscorrectscale}{%
\centering%
\fbox{\includegraphics%
[width=\expptscorrectscale,trim=360 250 200 300,clip]%
{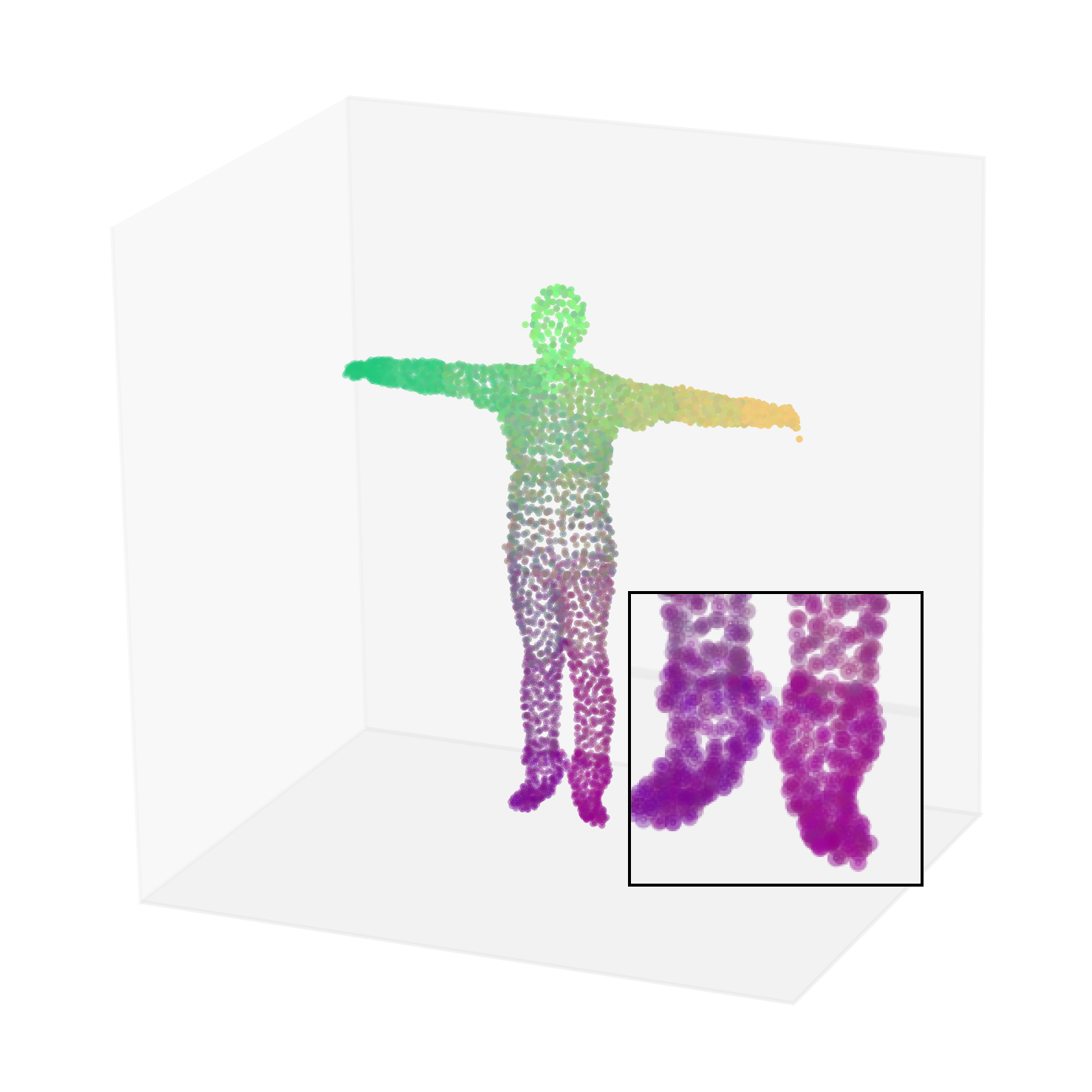}%
}\\%
{\footnotesize NPC optimized}%
}%
\centering%
\caption{\ourapproach{} recovers fine-grained geometry from the noisy initialization obtained by \cite{su2022danbo}.}%
\label{fig:supp-pts-correct}
\end{figure}
\begin{table}[t]
\caption{\textbf{ZJU-Mocap~\cite{peng2020neuralbody} subject 387 novel view and pose synthesis.} \ourapproach{} achieves improved FID and KID.}
\centering
\resizebox{1.0\linewidth}{!}{
\setlength{\tabcolsep}{3pt}
\begin{tabular}{lcccccccc}
\toprule%
& \multicolumn{4}{c}{ARAH} & \multicolumn{4}{c}{\ourapproach{} (Ours)} \\
\cmidrule(lr){2-5}\cmidrule(lr){6-9}%
 & PSNR  & LPIPS & FID & KID$\times$100  & PSNR~$\uparrow$  & LPIPS~$\downarrow$ & FID~$\downarrow$ & KID$\times100$~$\downarrow$ \\
\midrule%
Novel view & \textbf{25.83}& 0.096& 38.2& 1.17& 25.10& \textbf{0.093}& \textbf{36.1}& \textbf{0.94}\\%
\rowcolor{Gray}%
Novel pose & \textbf{22.93}& \textbf{0.127}& 51.4& 2.74& 21.88& 0.134& \textbf{49.2}& \textbf{2.59}\\%
\end{tabular}%
\label{tab:supp-zju}
}
\end{table}
To provide a further comparison on pose-dependent deformation, we tested on subject 387 of ZJU-Mocap~\cite{fang2021mirrored,peng2020neuralbody}, which includes loose clothing and long-range deformation dependency. We report the results in \tabref{tab:supp-zju} following the established protocols from~\cite{peng2020neuralbody}. \ourapproach{} match or outperform ARAH~\cite{wang2022arah} in perceptual metrics (LPIPS, KID, FID)
on novel-view and KID \& FID on novel-pose, despite ARAH using a prior pre-trained on a large-scale scan dataset~\cite{wang2021metaavatar}. A closer analysis (see \figref{fig:supp-zju}) reveals that texture %
details are improved, while pose-dependent wrinkle details are comparable if not better. ARAH yields overly smooth while \ourapproach{} produces slightly grainy results, leading to a lower PSNR but improved FID and KID. %

\subsection{\ourapproach~on Animal}
We verify the applicability of~\ourapproach~on entities where no pretrained surface prior available on the TAVA wolf cub subject~\cite{li2022tava}. We lightly train~\ourapproach~for 10k iterations, and present an out-of-distribution pose rendering in~\figref{fig:exp-animal-concept}.

\subsection{Ablation Study}
We use Human3.6M S9 to conduct our ablation study. In ~\tabref{tab:exp-ablation}, we show that all our proposed features contribute to the final perceptual quality. Without $f^\theta$, the rendering outcomes become noisy. Without $f^p$, the model produces inconsistent and blurry texture. Disabling $\Delta p$ results in lower perceptual quality for poses with more extreme deformation, as the nearest neighbor mapping becomes inaccurate in this case. In~\tabref{tab:supp-num-points}, we observe that \ourapproach{} is insensitive to the number of points, and using more than $200$ points (our default value) yields diminished returns. We further verify that \ourapproach{} learns character models with consistent quality by running our training framework 4 times, including re-training DANBO from scratch for canonical point clouds initialization. As reported in~\tabref{tab:supp-ablation-robust}, we observe low variation in the performance across different runs on both Human3.6M S9 and MonoPerfCap weipeng.

\newlength\expzjuscale
\setlength\expzjuscale{0.27\linewidth}
\ifblurface
\newcommand{\expzjupost}{_blur}
\else
\newcommand{\expzjupost}{}
\fi
\newcommand{\expzjupath}{statics/figs/supp/zju/}
\begin{figure}[t]
\setlength{\fboxsep}{0pt}%
\setlength{\fboxrule}{0pt}%
\parbox[t]{\expzjuscale}{%
\centering%
\fbox{\includegraphics%
[width=\expzjuscale,trim=00 0 00 20,clip]%
{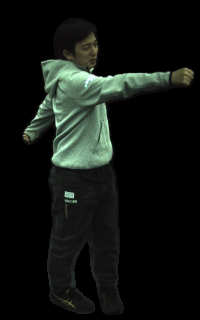}%
}\\%
{\scriptsize Training example}%
}%
\hfill
\parbox[t]{\expzjuscale}{%
\centering%
\fbox{\includegraphics%
[width=\expzjuscale,trim=00 0 00 20,clip]%
{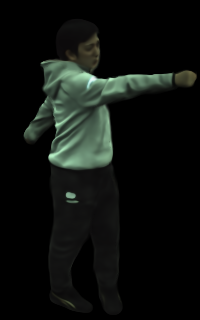}%
}\\%
{\footnotesize ARAH}%
}%
\hfill%
\parbox[t]{\expzjuscale}{%
\centering%
\fbox{\includegraphics%
[width=\expzjuscale,trim=00 0 00 20,clip]%
{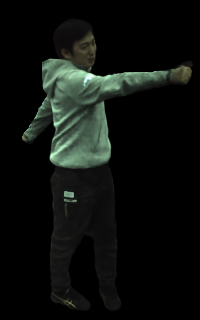}%
}\\%
{\footnotesize NPC (Ours)}%
}%
\centering%
\caption{\textbf{Novel view synthesis results on ZJU-Mocap subject 387.} Compared to ARAH,~\ourapproach~captures sharper but slightly grainy results.
}%
\label{fig:supp-zju}
\end{figure}
\newlength\expanimaltscale
\setlength\expanimaltscale{0.27\linewidth}
\newcommand{\expanimaltpath}{statics/figs/exp/animal/}

\begin{figure}[t]
\setlength{\fboxsep}{0pt}%
\setlength{\fboxrule}{0pt}%
\parbox[t]{\expanimaltscale}{%
\centering%
\fbox{\includegraphics%
[width=\expanimaltscale,trim=80 80 80 80,clip]%
{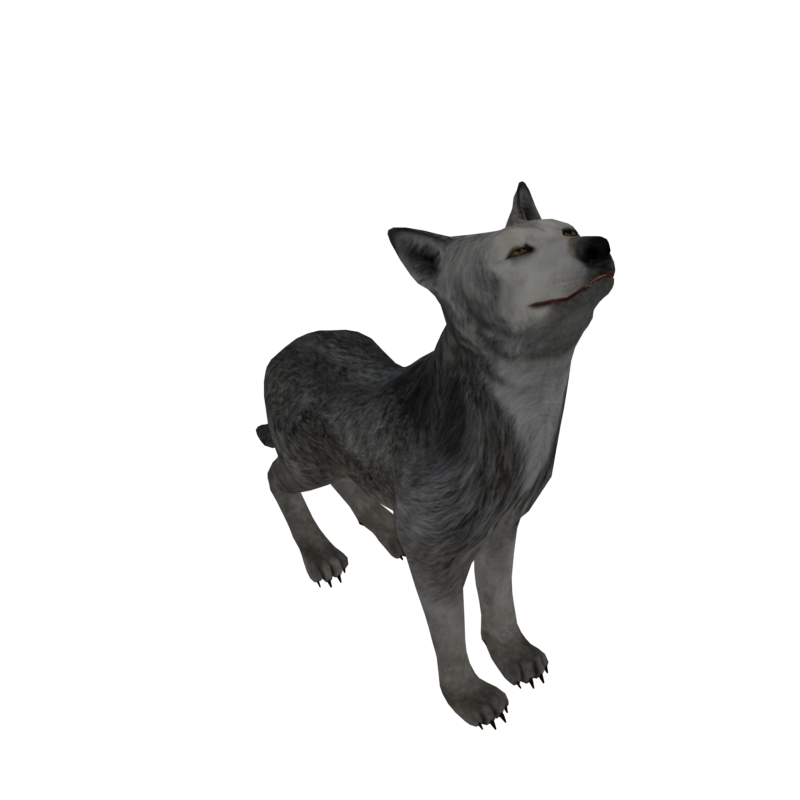}%
}\\%
{ Ground truth}\\%
}%
\hfill%
\parbox[t]{\expanimaltscale}{%
\centering%
\fbox{\includegraphics%
[width=\expanimaltscale,trim=0 0 0 0,clip]%
{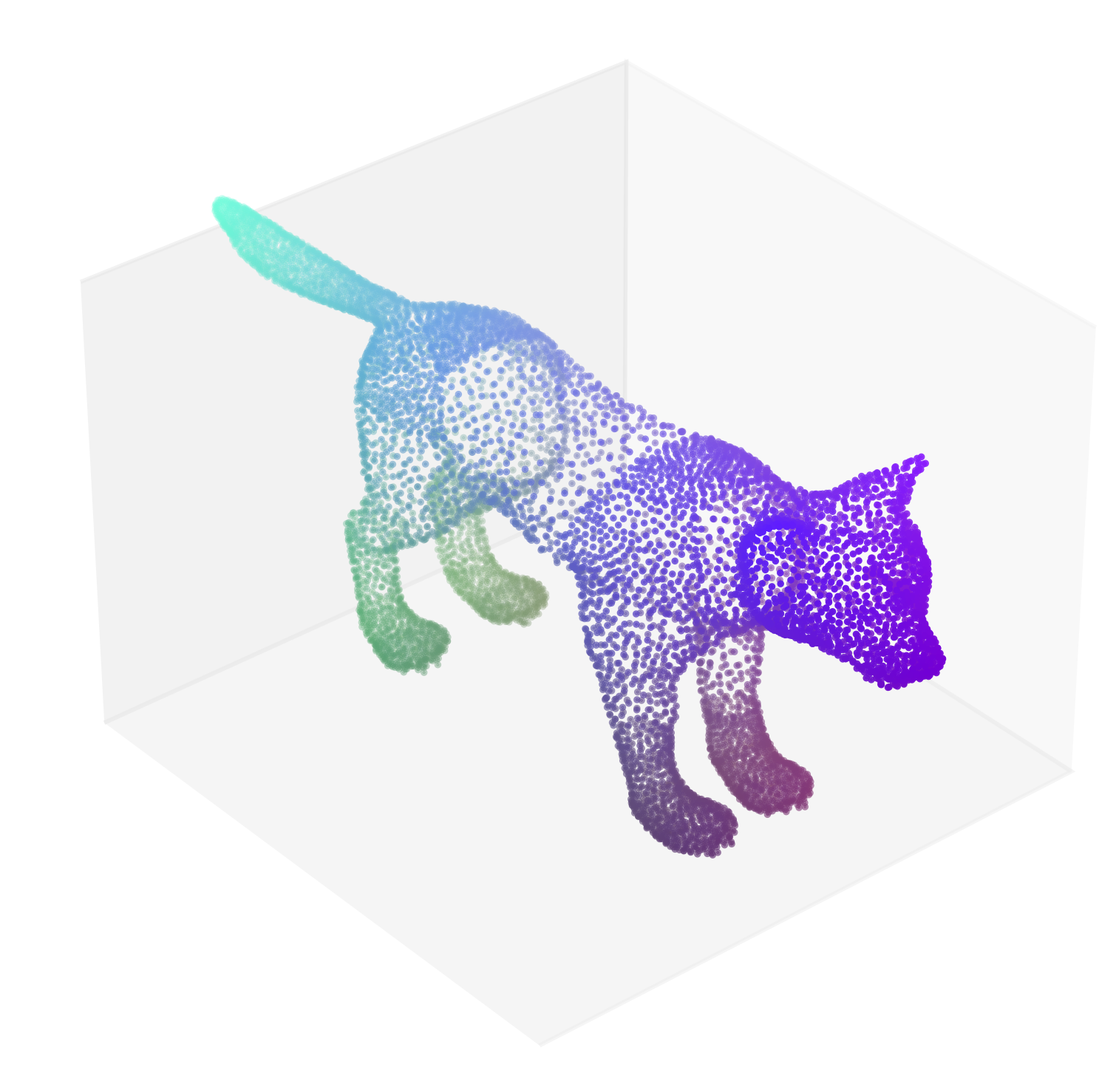}%
}\\%
{ Initial points}%
}%
\hfill%
\parbox[t]{\expanimaltscale}{%
\centering%
\fbox{\includegraphics%
[width=\expanimaltscale,trim=80 80 80 80,clip]%
{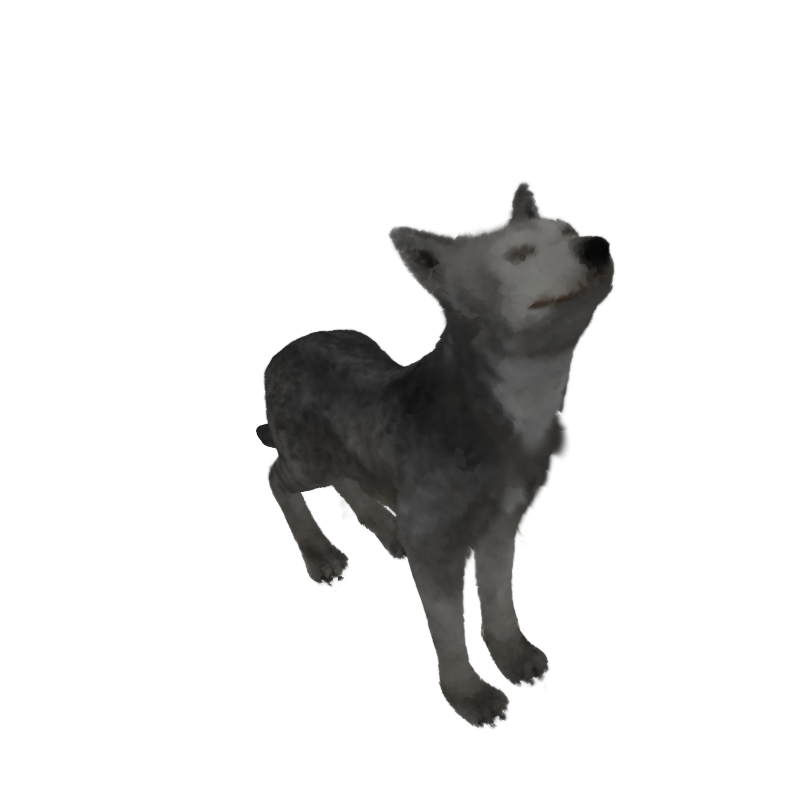}%
}\\%
{ NPC}%
}%
\centering%
\caption{\ourapproach~can be used for learning non-human subjects as it does not require a pre-trained surface prior.
}%
\label{fig:exp-animal-concept}
\end{figure}
\begin{table}[t]
\caption{\textbf{Ablation study on each of our proposed designs.} All of them contribute to the final perceptual quality.}
\centering
\resizebox{0.5\linewidth}{!}{
\setlength{\tabcolsep}{3pt}
\begin{tabular}{lcccc}
\toprule
 & KIDx100~$\downarrow$  & LPIPS~$\downarrow$ \\
\midrule
No $f^d$ & 5.54& 0.122\\
\rowcolor{Gray}
No $f^p$& 4.51& 0.122\\
No $f^\theta$& 4.50& 0.120\\
\rowcolor{Gray}
No $\Delta p$& 4.61& 0.120\\
\textbf{\ourapproach~(Ours)}& \textbf{4.43}& \textbf{0.115}\\
\end{tabular}
\label{tab:exp-ablation}
}
\end{table}
\begin{table}[t]
\caption{\textbf{Ablation study on the numbers of points per body parts}. \ourapproach{} shows consistent results with different numbers of points on Human 3.6M~\cite{Ionescu11}.
}
\centering
\resizebox{1.0\linewidth}{!}{
\setlength{\tabcolsep}{3pt}
\begin{tabular}{cccccccccccc}%
\toprule%
 \multicolumn{2}{c}{40$\times$19} & \multicolumn{2}{c}{120$\times$19} & \multicolumn{2}{c}{\textbf{200$\times$19}} & \multicolumn{2}{c}{280$\times$19} & \multicolumn{2}{c}{360$\times$19} \\%
\cmidrule(lr){1-2}\cmidrule(lr){3-4}\cmidrule(lr){5-6}\cmidrule(lr){7-8}\cmidrule(lr){9-10}%
   PSNR~$\uparrow$  & LPIPS~$\downarrow$& PSNR~$\uparrow$  & LPIPS~$\downarrow$  & PSNR~$\uparrow$   & LPIPS~$\downarrow$  & PSNR~$\uparrow$  &  LPIPS~$\downarrow$  & PSNR~$\uparrow$ & LPIPS~$\downarrow$ \\%
\midrule%
 \rowcolor{Gray}
24.83 & 0.123 & \textbf{24.88}&  0.117& 24.86& \underline{0.115}& \underline{24.87} & \textbf{0.114}& 24.84&  0.117\\%
\bottomrule%
\end{tabular}%
\label{tab:supp-num-points}
}
\end{table}
\begin{table}[t]
\caption{\textbf{Ablation study on running with 4 different canonical point clouds initializations}. We report the standard deviations in the parenthesis. The indicates that our point initialization strategy is reliable, and ~\ourapproach{} behaves consistently across different training runs.
}
\centering
\resizebox{1.0\linewidth}{!}{
\setlength{\tabcolsep}{3pt}
\begin{tabular}{ccccccccc}
\toprule
 \multicolumn{3}{c}{S9} & \multicolumn{3}{c}{WP} \\
\cmidrule(lr){1-3}\cmidrule(lr){4-6}%
  KIDx100~$\downarrow$  & LPIPS (VGG)~$\downarrow$  & LPIPS (Alex)~$\downarrow$ & KIDx100~$\downarrow$  & LPIPS (VGG)~$\downarrow$  & LPIPS (Alex)~$\downarrow$ \\
\midrule%
\rowcolor{Gray}
 4.34 ($\pm$ 0.12)& 0.116 ($\pm$ 0.000)& 0.124 ($\pm$ 0.000) & 3.75 ($\pm$ 0.29)& 0.207 ($\pm$ 0.001) & 0.127 ($\pm$ 0.001)\\
\end{tabular}
\label{tab:supp-ablation-robust}
}
\end{table}
\section{Limitations and Discussion}
Although \ourapproach{} improves runtime over most existing implicit neural body models, it still 
utilizes neural fields encoded by moderately large networks, 
which precludes real-time applications. 
Our approach sometimes produces small-scale ball-shaped artifacts when the sparse points cannot 
fully cover the body surfaces. This can potentially be resolved via point-growing as
proposed in Point-NeRF~\cite{xu2022point}. See the supplementary document for details.
Finally,~\ourapproach{} learns a person-specific model with detailed appearance 
from raw video sequences, providing a more accessible way for the public to create 
3D characters. 
On the downside, it could be exploited for DeepFakes, causing ethical concerns when used with malicious intent.
\section{Conclusion}
NPC demonstrated that detailed character models can be learned without having access 
to laser scanning technology or making restrictive assumptions on an underlying 
template model. 
As it is fully automatic, it makes digital full-body avatars available to a larger audience, 
including minorities that are not well-represented in existing datasets.

\ifarxiv
\paragraph{Acknowledgements.}
We thank Shaofei Wang and Ruilong Li for helpful discussions related to ARAH and TAVA. We thank Luis A. Bolaños for his help and discussions, and Frank Yu, Chunjin Song, Xingzhe He and Eric Hedlin for their insightful feedback. We also thank ARC at UBC and Compute Canada for providing computational resources.
\fi

{\small
\bibliographystyle{ieee_fullname}
\bibliography{bib/nerfhuman}
}

\end{document}